\documentclass{article}
\usepackage[nonatbib, final]{neurips_2024}

\usepackage{neurips_2024}

\usepackage[utf8]{inputenc} 
\usepackage[T1]{fontenc}    
\usepackage{hyperref}       
\usepackage{url}            
\usepackage{booktabs}       
\usepackage{amsfonts}       
\usepackage{nicefrac}       
\usepackage{microtype}      
\usepackage{xcolor}         

\usepackage[american]{babel}
\usepackage{mathtools}
\usepackage{tikz}

\usepackage{amsmath}
\usepackage{amssymb}
\usepackage{amsthm}
\usepackage{algorithm}
\usepackage[noend]{algpseudocode}

\newtheorem{assumption}{Assumption}

\newtheorem{corollary}{Corollary}
\newtheorem{theorem}{Theorem}

\newtheorem{lemma}{Lemma}
\newtheorem{remark}{Remark}
\newcommand{\citep}[1]{\cite{#1}}
\usepackage{footnote}
\makesavenoteenv{tabular}
\makesavenoteenv{table}

\title{Sample-Efficient Constrained Reinforcement Learning with General Parameterization}

\author{
    Washim Uddin Mondal\\
    Department of Electrical Engineering\\
    Indian Institute of Technology Kanpur\\
    Kanpur, UP, India 208016\\
    \texttt{wmondal@iitk.ac.in}\\
    \And
    Vaneet Aggarwal\\
    School of IE and ECE\\
    Purdue University\\
    West Lafayette, IN, USA 47906\\
    \texttt{vaneet@purdue.edu}
    }

\begin{document}
\maketitle
\begin{abstract}
We consider a constrained Markov Decision Problem (CMDP) where the goal of an agent is to maximize the expected discounted sum of rewards over an infinite horizon while ensuring that the expected discounted sum of costs exceeds a certain threshold. Building on the idea of momentum-based acceleration, we develop the Primal-Dual Accelerated Natural Policy Gradient (PD-ANPG) algorithm that ensures an $\epsilon$ global optimality gap and $\epsilon$ constraint violation with $\tilde{\mathcal{O}}((1-\gamma)^{-7}\epsilon^{-2})$ sample complexity for general parameterized policies where $\gamma$ denotes the discount factor. This improves the state-of-the-art sample complexity in general parameterized CMDPs by a factor of $\mathcal{O}((1-\gamma)^{-1}\epsilon^{-2})$ and achieves the theoretical lower bound in $\epsilon^{-1}$. 
\end{abstract}

\section{Introduction}
\label{sec:intro}

Reinforcement learning (RL) is a framework where an agent repeatedly interacts with an unknown Markovian environment to find a policy that maximizes the expected discounted sum of its observed rewards. Such problems, often modeled via Markov Decision Processes (MDPs), find applications in many areas, including transportation \citep{al2019deeppool}, communication networks \cite{geng2020multi}, robotics \cite{gonzalez2023asap}, etc. In many applications, however, the agents must also obey certain constraints. For example, in a food delivery network, the orders must be delivered within a stipulated time window; the marketing decisions of a firm must satisfy its budget constraints, etc. Such constraints are incorporated into RL by introducing a cost function. In constrained MDPs (CMDPs), the agents not only maximize the expected sum of discounted rewards but also ensure that the expected sum of discounted costs does not cross a predefined boundary.

Finding an optimal policy for a CMDP is a challenging problem, especially when the environment, i.e., the state transition function, is unknown. The efficiency of a solution to a CMDP is measured by its sample complexity, which essentially states how many state-transition samples it takes to yield a policy that is $\epsilon$ close to the optimal one while also ensuring that the expected sum of discounted costs does not violate the imposed boundary by more than $\epsilon$ amount. Many articles in the literature solve the CMDP with an unknown environment. Most of these works, however, focus on the tabular case where the number of states is finite. These solutions cannot be applied to many real-life scenarios where the state space is either large or infinite. To tackle this issue, the concept of policy parameterization must be invoked. Unfortunately, as exhibited in Table \ref{tab:my_label}, only a few works are available on CMDPs with parameterized policies. While, for softmax parameterization, the state-of-the-art (SOTA) sample complexity  is $\mathcal{O}(\epsilon^{-2})$, the same for the general parameterization is $\tilde{\mathcal{O}}(\epsilon^{-4})$ which is far from the lower bound $\Omega(\epsilon^{-2})$. It should be noted that the number of parameters needed in softmax parameterization is $\mathcal{O} (SA)$ where $S, A$ are the sizes of the state and action spaces of the underlying CMDP. On the other hand, general parameterization uses $\mathrm{d}\ll SA$ number of parameters, which makes it appropriate for dealing with large or infinite states. Given the importance of general parameterization for large state space CMDPs, the following question naturally arises: \textit{"Is it possible to solve CMDPs with general parameterization and achieve a sample complexity better than the SOTA $\tilde{\mathcal{O}}(\epsilon^{-4})$ bound?"}

In this article, we provide an affirmative answer to the above question. We propose a Primal-Dual-based Accelerated Natural Policy Gradient (PD-ANPG) algorithm to solve $\gamma$-discounted CMDPs with general parameterization. We theoretically prove that PD-ANPG achieves $\epsilon$ optimality gap and $\epsilon$ constraint violation with $\tilde{\mathcal{O}}((1-\gamma)^{-7}\epsilon^{-2})$ sample complexity (Theorem \ref{theorem_1}) that improves the SOTA $\tilde{\mathcal{O}}((1-\gamma)^{-8}\epsilon^{-4})$ sample complexity result of \citep{bai2023achieving}. It closes the gap between the theoretical upper and lower bounds of sample complexity in general parameterized CMDPs (in terms of $\epsilon^{-1}$), which was an open problem for quite some time (see the results of \cite{ding2020natural}, \cite{xu2021crpo}, \cite{bai2023achieving} in Table \ref{tab:my_label}). 

\subsection{Challenges and Key Insights}
\label{sub-sec:contribution}

Our algorithm builds upon the idea of primal dual-based NPG \citep{liu2020improved, bai2023achieving}. However, unlike the previous works, we use accelerated stochastic gradient descent (ASGD) in the inner loop to compute the estimate of the NPG. The improvement in the sample complexity results from two key observations. Firstly, we establish a global-to-local convergence lemma (Lemma \ref{lemma_lagrange_convergence}), which dictates how the global convergence of the Lagrange function is related to the first and second-order estimation error of the NPG. Here, via careful analysis, we show that the first-order term can be written as the expected bias of the NPG estimator (i.e., the difference between the true NPG and the expectation of its estimate). Secondly, we show (Lemma \ref{lemma_second_order} and its subsequent discussion) that the bias of the NPG estimate can be interpreted as the convergence error of an ASGD program with non-stochastic (i.e., deterministic) gradients. These, combined with the ASGD convergence result provided by \citep{jain2018accelerating}, lead to a convergence result of the Lagrange function (Corollary \ref{corr_lagrange_global_convergence}).

Finally, Theorem \ref{theorem_1} segregates the objective and constraint violation rates from  Lagrange convergence. Corollary \ref{corr_lagrange_global_convergence} shows that Lagrange convergence error is bounded by an independent function of $\zeta$ (the dual learning rate). One might, hence, be tempted to make $\zeta$ arbitrarily small. However, our analysis shows that although small $\zeta$ leads to better objective convergence, it worsens the constraint violation rate. We demonstrate how to optimally choose $\zeta$ to reach the middle ground which eventually leads us to $\tilde{\mathcal{O}}(\epsilon^{-2})$ sample complexity.

\subsection{Related Works}
\label{sub-sec:related-works}

\begin{table}[]
    \centering
    \begin{tabular}{|c|c|c|}
        \hline
         Algorithm & Sample Complexity & Parameterization \\ 
         \hline
         & & \\[-0.32cm]
         PMD-PD \citep{liu2021policy} & $\mathcal{O}(\epsilon^{-3})$ & Softmax\\
         \hline
         & & \\[-0.32cm]
         PD-NAC \citep{zeng2022finite} & $\mathcal{O}(\epsilon^{-6})$ & Softmax\\
         \hline
         & & \\[-0.32cm]
         NPG-PD \citep{ding2020natural} & $\mathcal{O}((1-\gamma)^{-5}\epsilon^{-2})$ & Softmax\\
         \hline
         & & \\[-0.32cm]
         CRPO \citep{xu2021crpo} & $\mathcal{O}((1-\gamma)^{-7}\epsilon^{-4})$ & Softmax\\
         \hline
         & & \\[-0.32cm]
         NPG-PD \citep{ding2020natural} & $\mathcal{O}((1-\gamma)^{-8}\epsilon^{-6})$ & General\\
         \hline
         & & \\[-0.32cm]
         CRPO \citep{xu2021crpo} & $\mathcal{O}((1-\gamma)^{-13}\epsilon^{-6})$ & General\\
         \hline
         & & \\[-0.32cm]
         C-NPG-PDA \citep{bai2023achieving} & $\tilde{\mathcal{O}}((1-\gamma)^{-8}\epsilon^{-4})$\footnote{We would like to point out that the sample complexity of C-NPG-PDA reported in \citep{bai2023achieving} is $\tilde{\mathcal{O}}((1-\gamma)^{-6}\epsilon^{-4})$. However, the result is erroneous. The authors have subsequently corrected their result, and the sample complexity has been modified to $\tilde{\mathcal{O}}((1-\gamma)^{-8}\epsilon^{-4})$ in the arXiv version (updated May 2024).} & General\\
         \hline
         & &\\[-0.32cm]
         \textbf{PD-ANPG (This Work)} & $\tilde{\mathcal{O}}((1-\gamma)^{-7}\epsilon^{-2})$ & General\\
         \hline
         & & \\[-0.35cm]
         Lower Bound \citep{vaswani2022near} & $\Omega((1-\gamma)^{-5}\epsilon^{-2})$ & $-$\\
         \hline
    \end{tabular}\vspace{0.1cm}
    \caption{Summary of sample complexity results on CMDP with parameterized policies. The parameter $\gamma$ indicates the discount factor. The dependence of the sample complexities of PMD-PD and PD-NAC on $\gamma$ is not depicted in \citep{liu2021policy,zeng2022finite}.  }
    \label{tab:my_label}
    \vspace{-.2in}
\end{table}

\textbf{Unconstrained RL:} 
Many algorithms solve MDPs with exact gradients \citep{agarwal2021theory, bhandari2021linear, cen2022fast, lan2023policy, zhan2023policy}. Moreover, many works use generative models to show either first-order \citep{xu2020improved, gargiani2022page, huang2020momentum, salehkaleybar2022adaptive, shen2019hessian} or global convergence \citep{chen2022finite, chen2022sample, khodadadian2022finite, fatkhullin2023stochastic, liu2020improved, masiha2022stochastic, mondal2024improved}.

\textbf{Constrained RL:} 
The tabular setting is well investigated, and many model-based \citep{efroni2020exploration, liu2021learning, ding2021provably, he2021nearly} and model-free \citep{xu2021crpo, ding2021provably, wei2021provably, bai2022achieving} algorithms are available in the literature. In comparison, there are relatively fewer works on parameterized policies. Policy mirror descent-primal dual (PMD-PD) algorithm was proposed by \citep{liu2021policy} that achieves $\mathcal{O}(\epsilon^{-3})$ sample complexity for softmax policies. For the same parameterization, \citep{zeng2022finite} achieved $\mathcal{O}(\epsilon^{-6})$ sample complexity via their proposed Online Primal-Dual Natural Actor-Critic Algorithm. The primal-dual Natural Policy Gradient algorithm suggested by \citep{ding2020natural} yields $\mathcal{O}(\epsilon^{-2})$ and $\mathcal{O}(\epsilon^{-6})$ sample complexities for softmax and general parametrization respectively.  \citep{xu2021crpo} also proposed a primal policy-based algorithm that works for both the softmax and general function approximation cases. The state of the art for the general parameterization is given by \citep{bai2023achieving}, where the $\tilde{\mathcal{O}}(\epsilon^{-4})$ sample complexity is obtained. This work improves upon this direction to obtain $\tilde{\mathcal{O}}(\epsilon^{-2})$ sample complexity. The comparisons are summarized in Table \ref{tab:my_label}. 

\section{Formulation}
\label{sec:formulation}
Let us consider a constrained Markov Decision Process (CMDP)  characterized by the tuple $\mathcal{M}=(\mathcal{S}, \mathcal{A}, r, c, P, \gamma, \rho)$ where $\mathcal{S}, \mathcal{A}$ denote the state space and the action space respectively, $r:\mathcal{S}\times \mathcal{A}\rightarrow [0, 1]$ defines the reward function, $c:\mathcal{S}\times\mathcal{A}\rightarrow [-1, 1]$ is the cost function, $P:\mathcal{S}\times\mathcal{A}\rightarrow \Delta(\mathcal{S})$ indicates the state transition kernel (where $\Delta(\mathcal{S})$ is the collection of all probability distributions over $\mathcal{S}$), $\gamma$ is the discount factor, and $\rho\in\Delta(\mathcal{S})$ is the initial state distribution. Note that the state space $\mathcal{S}$, in our setting, can potentially be a compact set of infinite size. However, for simplicity, we assume it to be countable. The action space, $\mathcal{A}$, is assumed to be of finite size. The range of the cost function is chosen to be $[-1, 1]$, rather than $[0, 1]$, to ensure that the constraint in our central optimization problem (defined later in \eqref{eq:original_optimization}) is non-trivial. A policy, $\pi:\mathcal{S}\rightarrow \Delta(\mathcal{A})$ is defined as a distribution over the action space for a given state of the environment. For a given policy, $\pi$, and a state-action pair $(s, a)$, we define the $Q$-value associated with $g\in\{r, c\}$ as follows.
\begin{align*}
    Q^{\pi}_g(s, a) \triangleq \mathbf{E}_{\pi}\left[\sum_{t=0}^\infty \gamma^t g(s_t, a_t)\bigg| s_0=s, a_0=a\right]
\end{align*}
where $\mathbf{E}_\pi$ is the expectation computed over all $\pi$-induced trajectories $\{(s_t, a_t)\}_{t=0}^\infty$ where $s_{t+1}\sim P(s_t, a_t)$ and $a_t\sim \pi(s_t)$, $\forall t\in \{0, 1, \cdots\}$. Similarly, the $V$-value associated with policy $\pi$, state $s$, and $g\in\{r, c\}$ is defined below.
\begin{align*}
    V^{\pi}_g(s) &\triangleq \mathbf{E}_{\pi}\left[\sum_{t=0}^\infty \gamma^t g(s_t, a_t)\bigg| s_0=s\right] 
    = \sum_{a}\pi(a|s)Q^{\pi}_g(s, a)
\end{align*}
Below we define the advantage value for a policy $\pi$, a state-action pair $(s, a)$, and $g\in \{r, c\}$. 
\begin{align*}
    A^{\pi}_g(s, a) \triangleq Q^{\pi}_g(s, a) - V^{\pi}_g(s)
\end{align*}
Define a function $J^{\pi}_{g, \rho}$, $\forall g\in\{r, c\}$ as follows.
\begin{align*}
    J^{\pi}_{g, \rho} \triangleq \mathbf{E}_{s\sim \rho} [V^{\pi}_g(s)]=\dfrac{1}{1-\gamma}\sum_{s, a}d^{\pi}_\rho(s) \pi(a|s)g(s, a)
\end{align*}
where $d_\rho^{\pi}\in\Delta(\mathcal{S})$ is the state occupancy measure given by,
\begin{align*}
    d^{\pi}_\rho(s) = (1-\gamma) \sum_{t=0}^\infty \gamma^t \mathrm{Pr}(s_t=s|s_0\sim \rho, \pi), ~\forall s\in \mathcal{S}
\end{align*}
Similarly, the state-action occupancy measure is defined as,
\begin{align}
    \label{eq:def_state_action_occupancy}
    \nu^{\pi}_\rho (s, a) = d^{\pi}_\rho(s)\pi(a|s), ~\forall (s, a)\in \mathcal{S} \times \mathcal{A}
\end{align}
Our goal is to maximize the function $J^{\pi}_{r,\rho}$ over all policies $\pi$ while ensuring that $J^{\pi}_{c, \rho}$ does not lie below a predefined threshold. Without loss of generality, we can formally express this problem as,
\begin{align}
    \begin{split}
        \max_{\pi}& ~J^{\pi}_{r, \rho} ~~~\text{subject to:} ~J^{\pi}_{c, \rho}\geq 0
    \end{split}
    \label{eq:original_optimization}
\end{align}
If the state space, $\mathcal{S}$, is large or infinite (which is the case in many application scenarios), the policies can no longer be represented in the tabular format; rather, they are indexed by a parameter, $\theta\in \Theta$. In this paper, we assume $\Theta = \mathbb{R}^{\mathrm{d}}$. Such indexing can be done via, for example, neural networks (NNs). Let $J_{g, \rho}(\theta)\triangleq J^{\pi_{\theta}}_{g, \rho}$. This allows us to redefine the constrained optimization problem as follows.
\begin{align}
    \begin{split}
        \max_{\theta\in\Theta}& ~J_{r, \rho}(\theta)~~~\text{subject to:} ~J_{c, \rho}(\theta)\geq 0
    \end{split}
    \label{eq:parameterized_optimization}
\end{align}
We assume the existence of at least one interior point solution of the above optimization. This is also known as Slater condition which can be formally expressed as follows.
\begin{assumption}
    \label{ass_slater}
    There exists $\Bar{\theta}$ such that $J_{c, \rho}(\Bar{\theta})\geq c_{\mathrm{slater}}$ for some $c_{\mathrm{slater}}\in (0, \frac{1}{1-\gamma}]$.
\end{assumption}

\section{Algorithm}
\label{sec:algorithm}

The dual problem associated with the constraint optimization \eqref{eq:parameterized_optimization} can be written as follows.
\begin{align}
    \begin{split}
         \min_{\lambda\geq 0}\max_{\theta\in\Theta} ~& J_{\mathrm{L},\rho}(\theta, \lambda)~~
        \text{where}~ J_{\mathrm{L},\rho}(\theta, \lambda) \triangleq J_{r, \rho}(\theta) + \lambda J_{c, \rho}(\theta)
    \end{split}
\end{align}
The function, $J_{\mathrm{L}, \rho}(\cdot, \cdot)$ is called the Lagrange function while $\lambda$ is said to be the Lagrange multiplier.  The above problem can be solved by iteratively applying the following update rule $\forall k\in\{0, \cdots, K-1\}$, starting with $(\theta_0, \lambda_0)$ where $\theta_0\in\Theta$ is arbitrary and $\lambda_0=0$.
\begin{align}
    \label{eq:update_theta}
    \theta_{k+1} &= \theta_k + \eta F_{\rho}(\theta_k)^{\dagger}\nabla_{\theta} J_{\mathrm{L}, \rho}(\theta_k, \lambda_k)\\
    \label{eq:update_lambda}
    \lambda_{k+1} &= \mathcal{P}_{\Lambda}\left[\lambda_k - \zeta J_{c,\rho}(\theta_k)\right]
\end{align}
where $\eta, \zeta$ are learning rates, $\mathcal{P}_{\Lambda}$ denotes the projection function onto the set, $\Lambda\triangleq[0, \lambda_{\max}]$, and $\dagger$ is the Moore-Penrose pseudoinverse operator. The choice of $\lambda_{\max}$ will be specified later. Note that the update rule of $\theta$ is similar to that of the standard policy gradient method except here the learning rate, $\eta$ is modulated by the inverse of the Fisher matrix, $F_\rho(\theta)$ which is defined below. 
\begin{align}
    \label{eq_def_F_rho}
    F_\rho(\theta) \triangleq \mathbf{E}_{(s, a)\sim \nu^{\pi_\theta}_\rho}\left[\nabla_\theta \log \pi_\theta(a|s) \otimes \nabla_\theta \log \pi_\theta(a|s)\right]
\end{align}
where $\otimes$ indicates the outer product. Using a variation of the classical policy gradient theorem \citep{sutton1999policy}, one can obtain the gradient of the Lagrange function as follows.
\begin{align}
    \label{eq:policy_grad_theorem}
    \begin{split}
        \nabla_\theta J_{\mathrm{L}, \rho}(\theta, \lambda) = \dfrac{1}{1-\gamma} H_{\rho}(\theta, \lambda), ~\text{where}~ &H_\rho(\theta, \lambda) \triangleq \mathbf{E}_{(s, a)\sim \nu^{\pi_\theta}_\rho} \left[A^{\pi_\theta}_{\mathrm{L}, \lambda}(s, a)\nabla_\theta \log \pi_\theta (a|s) \right]\\
        ~\text{and}~ &A^{\pi_\theta}_{\mathrm{L}, \lambda}(s, a) \triangleq A^{\pi_\theta}_{r}(s, a) + \lambda A^{\pi_\theta}_{c}(s, a)
    \end{split}
\end{align}
In most application scenarios, the learner is unaware of the state transition function, $P$, and thereby, of the advantage function, $A^{\pi_\theta}_{\mathrm{L}, \lambda}$ and the occupancy measure, $\nu^{\pi_\theta}_\rho$. This makes the exact computation of $F_\rho(\theta)$ and $H_\rho(\theta, \lambda)$ an impossible task. Fortunately, there is a way to obtain an approximate value of the \textit{natural policy gradient} $\omega^*_{\theta, \lambda}\triangleq F_\rho(\theta)^{\dagger} \nabla_\theta J_{\mathrm{L}, \rho}(\theta, \lambda)$ that does not require the knowledge of $P$. Invoking \eqref{eq:policy_grad_theorem}, one can prove that $\omega_{\theta, \lambda}^*$ is a solution of a quadratic optimization. Formally, we have,
\begin{align}
    \label{eq_quad_optimization}
    \begin{split}
        \omega^*_{\theta, \lambda} &\in {\arg\min}_{\omega\in \mathbb{R}^{\mathrm{d}}} L_{\nu^{\pi_\theta}_\rho}(\omega, \theta, \lambda),\\
        \text{where}~L_{\nu^{\pi_\theta}_\rho}(\omega, \theta, \lambda) &\triangleq \dfrac{1}{2}\mathbf{E}_{(s, a)\sim \nu^{\pi_\theta}_\rho}\left[ \left(\dfrac{1}{1-\gamma} A^{\pi_\theta}_{\mathrm{L}, \lambda}(s, a) - \omega^{\mathrm{T}}\nabla_\theta \log\pi_\theta(a|s)\right)^2\right]
    \end{split}
\end{align}
The above reformulation opens up the possibility to compute $\omega^*_{\theta, \lambda}$ via a gradient descent-type iterative procedure. Observe that the gradient of $L_{\nu_\rho^{\pi_\theta}}(\cdot, \theta, \lambda)$ can be calculated as follows.
\begin{align}
    \label{eq:quadratic_gradient}
    \nabla_\omega L_{\nu^{\pi_\theta}_\rho} (\omega, \theta, \lambda) = F_\rho(\theta)\omega - \dfrac{1}{1-\gamma} H_\rho (\theta, \lambda)
\end{align}
 Algorithm \ref{algo_sampling} describes a procedure to obtain unbiased estimates of this gradient. This is inspired by Algorithm 3 of \cite{agarwal2021theory}. Additionally, observe from \eqref{eq:update_lambda} that the update of the Lagrange variable, $\lambda$ requires the computation of $J_{c,\rho}(\theta)$ which is also difficult to accomplish without having an explicit knowledge about $P$. Algorithm \ref{algo_sampling} also provides an unbiased estimation of the above quantity. 
\begin{algorithm}[t!]
    \begin{algorithmic}[1]
    \caption{Unbiased Sampling}
        \State \textbf{Input:} Parameters $(\theta,\omega, \lambda, \gamma)$, Initial Distribution $\rho$
        \vspace{0.2cm}
        \State $T\sim \mathrm{Geo}(1-\gamma)$, $s_0\sim \rho$, $a_0\sim \pi_{\theta}(s_0)$
        \For{$j\in\{0,\cdots, T-1\}$}
            \State Execute $a_{j}$, observe $s_{j+1}\sim P(s_j, a_j)$ and sample $a_{j+1}\sim \pi_{\theta}(s_{j+1})$
        \EndFor
        \State $\hat{J}_{c, \rho}(\theta) \gets \sum_{j=0}^T c(s_j, a_j)$, $(\hat{s}, \hat{a})\gets (s_{T}, a_{T})$
        \vspace{0.2cm}
        \State $T\sim \mathrm{Geo}(1-\gamma)$, $(s_0, a_0)\gets (\hat{s}, \hat{a})$ \Comment{Q-function Estimation}
        \For{$j\in\{0,\cdots, T-1\}$}
            \State Execute $a_{j}$, observe $s_{j+1}\sim P(s_j, a_j)$, and sample $a_{j+1}\sim \pi_{\theta}(s_{j+1})$
        \EndFor
        \For{$g\in \{r, c\}$}
            \State $\hat{Q}_g^{\pi_{\theta}}(\hat{s}, \hat{a})\gets \sum_{j=0}^{T}g(s_j, a_j)$
        \EndFor
        \vspace{0.2cm}
        \State $T\sim \mathrm{Geo}(1-\gamma)$, $s_0\gets\hat{s}$, $a_0\sim \pi_{\theta}(s_0)$ \Comment{V-function Estimation}
        \For{$j\in\{0,\cdots, T-1\}$}
            \State Execute $a_{j}$, observe $s_{j+1}\sim P(s_j, a_j)$, and sample $a_{j+1}\sim \pi_{\theta}(s_{j+1})$
        \EndFor
        \For{$g\in \{r, c\}$}
            \State $\hat{V}_g^{\pi_{\theta}}(\hat{s})\gets \sum_{j=0}^{T}g(s_j, a_j)$
            \State $\hat{A}_g^{\pi_{\theta}}(\hat{s}, \hat{a})\gets \hat{Q}_g^{\pi_{\theta}}(\hat{s}, \hat{a})-\hat{V}_g^{\pi_{\theta}}(\hat{s})$
        \EndFor
        \State \Comment{Estimation of Relevant functions}
        \begin{flalign}
        &\hat{A}^{\pi_\theta}_{\mathrm{L}, \lambda}(\hat{s}, \hat{a}) \gets \hat{A}^{\pi_\theta}_{r}(\hat{s}, \hat{a}) + \lambda \hat{A}^{\pi_\theta}_{c}(\hat{s}, \hat{a})\\
            \label{eq:estimate_F_rho}
            &\hat{F}_{\rho}(\theta)\gets \nabla_{\theta}\log\pi_{\theta}(\hat{a}|\hat{s})\otimes\nabla_{\theta}\log\pi_{\theta}(\hat{a}|\hat{s})\\
            \label{eq:estimate_H_rho}
            &\hat{H}_{\rho}(\theta, \lambda)\gets \hat{A}_{\mathrm{L}, \lambda}^{\pi_\theta}(\hat{s}, \hat{a})\nabla_{\theta}\log \pi_\theta(\hat{a}|\hat{s})&
        \end{flalign}
        \State  \Comment{Gradient Estimate}
        \begin{align}
            \label{eq:estimation_gradient}
            \hat{\nabla}_\omega L_{\nu^{\pi_\theta}_\rho}(\omega, \theta, \lambda)\gets \hat{F}_{\rho}(\theta)\omega -\dfrac{1}{1-\gamma}\hat{H}_{\rho}(\theta, \lambda)
        \end{align}
        \State \textbf{Output:} $\hat{J}_{c, \rho}(\theta), \hat{\nabla}_{\omega} L_{\nu^{\pi_\theta}_\rho}(\omega, \theta, \lambda)$
        \label{algo_sampling}
    \end{algorithmic}
\end{algorithm}

Algorithm \ref{algo_sampling} first samples a horizon length, $T$, from the geometric distribution with success probability $(1-\gamma)$ and executes the CMDP for $T$ instances following the policy, $\pi_\theta$, starting from a state $s_0\sim \rho$. The total cost observed in the resulting trajectory is assigned as the estimate $\hat{J}_{c, \rho}(\theta)$. The state-action pair $(s_T, a_T)$ can be assumed to be an arbitrary sample $(\hat{s}, \hat{a})$ chosen from the occupancy measure $\nu^{\pi_\theta}_\rho$. The algorithm then generates a $\pi_\theta$-induced trajectory of length $T\sim \mathrm{Geo}(1-\gamma)$, taking $(\hat{s}, \hat{a})$ as the starting point. The total reward and cost observed in this trajectory are assigned as $\hat{Q}_r^{\pi_\theta}(\hat{s}, \hat{a})$ and $\hat{Q}_c^{\pi_\theta}(\hat{s}, \hat{a})$ respectively. Next, another $\pi_\theta$-induced trajectory of length $T\sim \mathrm{Geo}(1-\gamma)$ is generated assuming the state, $\hat{s}$ as the initiation point. The total reward and cost of this trajectory are assigned as $\hat{V}^{\pi_\theta}_r(\hat{s})$ and $\hat{V}^{\pi_\theta}_c(\hat{s})$ respectively. For $g\in\{r, c\}$, an estimate of the advantage value is computed as $\hat{A}^{\pi_\theta}_g(\hat{s}, \hat{a})=\hat{Q}^{\pi_\theta}_g(\hat{s}, \hat{a})-\hat{V}^{\pi_\theta}_g(\hat{s})$. Finally, the estimates of $F_\rho(\theta)$ and $H_\rho(\theta, \lambda)$ are obtained via \eqref{eq:estimate_F_rho} and \eqref{eq:estimate_H_rho} respectively which produces an estimation of the desired gradient in \eqref{eq:estimation_gradient}. The following Lemma demonstrates that the estimates produced by Algorithm \ref{algo_sampling} are unbiased.
\begin{lemma}
\label{lemma_unbiased}
    Let $\hat{J}_{c, \rho}(\theta)$,  $\hat{\nabla}_{\omega}L_{\nu^{\pi_\theta}_\rho}(\omega, \theta, \lambda)$ be the estimates produced by Algorithm \ref{algo_sampling} for a predefined set of parameters $(\omega, \theta, \lambda)$. The following equations hold.
    \begin{align*}
        & \mathbf{E}\left[\hat{J}_{c, \rho}(\theta)\big|\theta\right] = J_{c,\rho}(\theta)~~\text{and}~~\mathbf{E}\left[\hat{\nabla}_{\omega}L_{\nu^{\pi_\theta}_\rho}(\omega, \theta, \lambda)\big| \omega, \theta, \lambda \right] = \nabla_{\omega}L_{\nu^{\pi_\theta}_\rho}(\omega, \theta, \lambda)
    \end{align*}
\end{lemma}
In the absence of knowledge about the transition model, $P$, one can utilize the estimates generated by Algorithm \ref{algo_sampling} as good proxies for their true values. In particular, one can obtain an approximate value of the natural policy gradient $\omega^*_{\theta, \lambda}$ by iteratively minimizing the function $L_{\nu_\rho^{\pi_\theta}}(\cdot, \theta, \lambda)$ using the gradient estimate $\hat{\nabla}_{\omega}L_{\nu^{\pi_\theta}_\rho}(\omega, \theta, \lambda)$. On the other hand, using the estimate $\hat{J}_{c, \rho}(\theta)$, an approximate update equation of the Lagrange parameter can be formed. Algorithm \ref{algo_npg} uses these two ideas to obtain a policy that is close to the optimal one.

Algorithm \ref{algo_npg} has a nested loop structure. The \textit{outer loop} runs $K$ number of times. At a given instance, $k$, of the outer loop, the policy parameter $\theta_k$ and the Lagrange parameter, $\lambda_k$ are updated via \eqref{eq:policy_par_update} and \eqref{eq:lambda_update}. The estimate, $\hat{J}_{c, \rho}(\theta_k)$ is computed via Algorithm \ref{algo_sampling}. On the other hand, $\omega_k$, the approximate value of the natural policy gradient $\omega_{\theta_k, \lambda_k}^*$ is obtained by iteratively minimizing $L_{\nu_\rho^{\pi_{\theta_k}}}(\cdot, \theta_k, \lambda_k)$ in $H$ number of \textit{inner loop} steps via the Accelerated Stochastic Gradient Descent (ASGD) procedure as stated in \citep{jain2018accelerating}. ASGD comprises the iterative updates \eqref{eq:asgd_1}$-$\eqref{eq:asgd_4} with tunable learning parameters $(\alpha, \beta, \xi, \delta)$ followed by a tail-averaging step \eqref{eq:tail_average}. The gradient estimate utilized in \eqref{eq:asgd_2} and \eqref{eq:asgd_4} is obtained via Algorithm \ref{algo_sampling}. It is worth mentioning that existing NPG algorithms such as that given in \citep{liu2020improved} typically apply the SGD, rather than the ASGD procedure, to obtain $\omega_k$. The difference between these subroutines is that while SGD uses only the current gradient estimate to update $\omega_k$, ASGD considers the contribution of all previous gradient estimates (momentum) using its convoluted iteration and tail-averaging steps. 

\begin{algorithm}[t!]
    \caption{Primal-Dual Accelerated Natural Policy Gradient (PD-ANPG)}
    \begin{algorithmic}[1]
        \State \textbf{Input:}  Parameters $(\theta_0, \lambda_0)$, Distribution $\rho$,
        Run-time Parameters $K, H$, Learning Parameters $\eta, \zeta, \alpha, \beta, \xi, \delta$
        \vspace{0.1cm}
        \For{$k\in\{0, \cdots, K-1\}$}
        \Comment{Outer Loop}
        \State $\mathbf{x}_0, \mathbf{v}_0\gets \mathbf{0}$
        \For{$h\in\{0, \cdots, H-1\}$} \Comment{Inner Loop}
        \State \Comment{Accelerated Stochastic Gradient Descent}
        \begin{align}
            \label{eq:asgd_1}
            & \mathbf{y}_h \gets \alpha\mathbf{x}_{h}+(1-\alpha)\mathbf{v}_h
        \end{align}
        \State $\hat{G}\gets\hat{\nabla}_{\omega} L_{\nu^{\pi_{\theta_k}}_\rho}(\omega,\theta_k, \lambda_k)\big|_{\omega=\mathbf{y}_h}$ (Algorithm \ref{algo_sampling})
        \begin{align}
        \label{eq:asgd_2}
            &\mathbf{x}_{h+1}\gets \mathbf{y}_h - \delta \hat{G}\\
            \label{eq:asgd_3}
            & \mathbf{z}_h \gets \beta \mathbf{y}_h + (1-\beta) \mathbf{v}_h\\
            \label{eq:asgd_4}
            & \mathbf{v}_{h+1}\gets \mathbf{z}_h - \xi \hat{G}
        \end{align}
        \EndFor
        \State Tail Averaging:
        \begin{align}
            \omega_k\gets \dfrac{2}{H}\sum_{\frac{H}{2}<h\leq H} \mathbf{x}_h
            \label{eq:tail_average}
        \end{align}
        \State Obtain $\hat{J}_{c, \rho}(\theta_k)$ via Algorithm \ref{algo_sampling}.
        \State Parameter Updates:
        \begin{align}
            \label{eq:policy_par_update}
            &\theta_{k+1}\gets \theta_k + \eta \omega_k\\
            \label{eq:lambda_update}
            &\lambda_{k+1}\gets \mathcal{P}_{\Lambda} [\lambda_k-\zeta \hat{J}_{c, \rho}(\theta_k)]
        \end{align}
        \EndFor
        \State \textbf{Output:} $\{\theta_k\}_{k=0}^{K-1}$
    \end{algorithmic}
    \label{algo_npg}
\end{algorithm}
 
\section{Analysis}
\label{sec:analysis}
Our goal in this section is to characterize the rate of convergence of the objective function and the constraint violation if policy parameters are generated via Algorithm \ref{algo_npg}. We start by stating a few assumptions needed for the analysis.
\begin{assumption}
    \label{ass_score}
    The log-likelihood function is $G$-Lipschitz and $B$-smooth where $B, G>0$. Formally, the following relations hold $\forall \theta, \theta_1,\theta_2 \in\Theta$, and $\forall (s,a)\in \mathcal{S}\times \mathcal{A}$.
    \begin{align*}
            \Vert \nabla_\theta\log\pi_\theta(a\vert s)\Vert&\leq G,~\text{and}~
            \Vert \nabla_\theta\log\pi_{\theta_1}(a\vert s)-\nabla_\theta\log\pi_{\theta_2}(a\vert s)\Vert\leq B\Vert \theta_1-\theta_2\Vert\quad
    \end{align*}
\end{assumption}

\begin{remark}
    Assumption \ref{ass_score} is commonly applied in proving convergence guarantees of policy gradient-type algorithms \citep{Mengdi2021, agarwal2021theory, liu2020improved}. This assumption is obeyed by many widely used policy classes such as the class of neural networks with bounded weights.
\end{remark}

Assumption \ref{ass_score} implies the boundedness of the gradient of the Lagrange function. This can be formally expressed as follows.
\begin{lemma}
\label{lemma_2}
    If Assumption \ref{ass_score} holds, then the following inequality is true $\forall \theta\in \Theta$ and $\forall \lambda\in \Lambda$.
    \begin{align*}
        \Vert\nabla_\theta J_{\mathrm{L}, \rho}(\theta, \lambda)\Vert\leq \dfrac{G(1+\lambda_{\max})}{(1-\gamma)^2}
    \end{align*}
\end{lemma}
\begin{proof}
    Statement (a) can be proven using \eqref{eq:policy_grad_theorem} along with Assumption \ref{ass_score} and the facts that $|A_{\mathrm{L}, \lambda}^{\pi_\theta}(s, a)|$ is bounded by $(1+\lambda)/(1-\gamma)$ and $\lambda\leq \lambda_{\max}$, $\forall \lambda \in \Lambda$. 
\end{proof}

The result established by Lemma \ref{lemma_2} will be pivotal in our further analysis.

\begin{assumption}
    \label{ass_transfer_error}
    The compatible function approximation error defined in \eqref{eq_quad_optimization} satisfies the inequality $L_{\nu_\rho^{\pi^*}}(\omega^*_{\theta, \lambda}, \theta, \lambda)\leq \epsilon_{\mathrm{bias}}/2$, $\forall \theta\in \Theta$ and $\forall \lambda\in \Lambda$ where $\pi^*$ is a solution to the original constrained optimization problem \eqref{eq:original_optimization} and $\omega_{\theta, \lambda}^*$ is defined in \eqref{eq_quad_optimization}. The term $\epsilon_{\mathrm{bias}}$ is a non-negative constant. The factor $2$ is used for notational convenience.
\end{assumption}

\begin{remark}
    The term $\epsilon_{\mathrm{bias}}$ quantifies the expressivity of the parameterized policy class. For example, if the parameterization is complete i.e., includes all possible policies (such as in direct or softmax parameterization), then $\epsilon_{\mathrm{bias}}=0$ \citep{agarwal2021theory}. A similar result can be proven for linear MDPs \citep{Chi2019}. For incomplete policy classes, we have $\epsilon_{\mathrm{bias}}>0$. However, if the class is sufficiently rich (such as neural networks with a large number of parameters), $\epsilon_{\mathrm{bias}}$ can be assumed to be negligibly small \citep{wang2019neural}.
\end{remark}

\begin{assumption}
    \label{ass_fisher}
    There exists a positive constant $\mu_F$ such that $F_\rho(\theta)-\mu_FI_{\mathrm{d}}$ is positive semidefinite i.e., $F_\rho(\theta)\succeq \mu_FI_{\mathrm{d}}$, $\forall \theta\in \Theta$ where $I_{d}$ is a $\mathrm{d}\times \mathrm{d}$ identity matrix and $F_\rho(\cdot)$ is defined in \eqref{eq_def_F_rho}.
\end{assumption}

The property of the policy classes laid out in Assumption \ref{ass_fisher} is called Fisher Non-Degeneracy (FND) which essentially ensures that $\forall \theta\in\Theta$, the Fisher matrix $F_\rho(\theta)$ is away from the zero matrix by a certain amount. Observe that the Hessian of the function, $l_{\theta, \lambda}(\cdot)\triangleq L_{\nu_\rho^{\pi_\theta}}(\cdot, \theta, \lambda)$ is $F_\rho(\theta)$. Therefore, Assumption \ref{ass_fisher} also indicates that $l_{\theta, \lambda}$ is $\mu_F$-strongly convex. This assumption is commonly applied in analyzing policy-gradient algorithms \citep{bai2023achieving, zhang2020global, liu2020improved}. Assumption \ref{ass_fisher} also ensures that the matrix $F_\rho(\theta)$ is invertible, which, in turn, implies the uniqueness of the maximizer $\omega_{\theta, \lambda}^*=\arg\min_{\omega\in\mathbb{R}^{\mathrm{d}}}l_{\theta, \lambda}(\omega)$. \citep{mondal2023mean} describes a concrete set of policies that obeys Assumption $\ref{ass_score}-\ref{ass_fisher}$.

\subsection{Local-to-Global Convergence Lemma}
Recall that our goal is to establish the global convergence rates. Lemma \ref{lemma_lagrange_convergence} (stated below) is the first step in that direction. Specifically, it demonstrates how the average optimality gap of the Lagrange function can be bounded by the first and second-order error of the gradient estimates.

\begin{lemma}
    \label{lemma_lagrange_convergence} 
    If the parameters $\{\theta_k, \lambda_k\}_{k=0}^{K-1}$ are updated via \eqref{eq:policy_par_update} and \eqref{eq:lambda_update} and assumptions $\ref{ass_score}-\ref{ass_fisher}$ hold, then the following inequality holds for any $K$.
    \begin{equation}
        \label{eq:general_bound}
	\begin{split}
            \frac{1}{K}\sum_{k=0}^{K-1}\mathbf{E}\bigg(J_{\mathrm{L}, \rho}(\pi^*, \lambda_k)&-J_{\mathrm{L}, \rho}(\theta_k,\lambda_k)\bigg)\leq \sqrt{\epsilon_{\mathrm{bias}}}+\frac{G}{K}\sum_{k=0}^{K-1}\mathbf{E}\Vert(\mathbf{E}\left[\omega_k\big|\theta_k, \lambda_k\right]-\omega^*_k)\Vert\\
            &+\frac{B\eta}{2K}\sum_{k=0}^{K-1}\mathbf{E}\Vert\omega_k\Vert^2
            +\frac{1}{\eta K}\mathbf{E}_{s\sim d^{\pi^*}_\rho}[KL(\pi^*(\cdot\vert s)\Vert\pi_{\theta_0}(\cdot\vert s))]		
        \end{split}
    \end{equation}
    where $\omega^*_k\triangleq\omega^*_{\theta_k, \lambda_k}$, $\omega^*_{\theta_k,\lambda_k}$ is the natural policy gradient defined in \eqref{eq_quad_optimization}, and $\pi^*$ is the solution to the constrained optimization $\eqref{eq:original_optimization}$. Finally, $\omega_k$ is the approximation of $\omega_k^*$ given by \eqref{eq:tail_average} and $KL(\cdot\Vert\cdot)$ is the KL-divergence.
\end{lemma}

Note the presence of the term, $\epsilon_{\mathrm{bias}}$ in \eqref{eq:general_bound}. It shows that due to the incompleteness of the parameterized policy class, the average optimality error cannot be made arbitrarily small. It is worth mentioning that many existing CMDP analyses (such as \citep{bai2023achieving}) follow a path similar to that of Lemma \ref{lemma_lagrange_convergence}. However, while the first order term in those works turns out to be $\mathbf{E}\Vert\omega_k-\omega_k^*\Vert$, we improved it to $\mathbf{E}\Vert\mathbf{E}[\omega_k|\theta_k, \lambda_k]-\omega_k^*\Vert$. Such seemingly insignificant improvement has important ramifications for our analysis as explained later in the paper. The second order term in \eqref{eq:general_bound} can be expanded as, 
\begin{align}
\label{eq_second_order_expand}
\begin{split}
    \dfrac{1}{K}\sum_{k=0}^{K-1} \mathbf{E}\Vert \omega_k\Vert^2 &\leq \dfrac{2}{K}\sum_{k=0}^{K-1}\mathbf{E}\Vert \omega_k-\omega_k^*\Vert^2 +\dfrac{2}{K}\sum_{k=0}^{K-1}\mathbf{E}\Vert\omega_k^*\Vert^2\\
    &\overset{(a)}{\leq} \dfrac{2}{K}\sum_{k=0}^{K-1}\mathbf{E}\Vert \omega_k-\omega_k^*\Vert^2 +\dfrac{2}{\mu_F^{2}K}\sum_{k=0}^{K-1}\mathbf{E}\left\Vert \nabla_\theta J_{\mathrm{L}, \rho}(\theta_k, \lambda_k)\right\Vert^2\\
    &\overset{(b)}{\leq} \dfrac{2}{K}\sum_{k=0}^{K-1}\mathbf{E}\Vert \omega_k-\omega_k^*\Vert^2 + \dfrac{2G^2(1+\lambda_{\max})^2}{\mu_F^2 (1-\gamma)^4}
\end{split}
\end{align}
where (a) utilises $\omega_k^*=F_\rho(\theta_k)^{\dagger}\nabla_{\theta}J_{\mathrm{L}, \rho}(\theta_k, \lambda_k)$ and Assumption \ref{ass_fisher}. The second inequality applies Lemma \ref{lemma_2} together with the fact that $\lambda_k\in \Lambda$. Our next subsection provides a bound on $\mathbf{E}\Vert \omega_k-\omega_k^*\Vert^2$ and the first order term $\mathbf{E}\Vert \mathbf{E}[\omega_k|\theta_k, \lambda_k] -\omega_k^*\Vert$.

\vspace{-.1in}
\subsection{Local Convergence of the Natural Policy Gradient}
\vspace{-.1in}
To deliver the promised bounds, we first provide some characterization of the gradient estimate. 

\begin{lemma}
\label{lemma_5}
Let $\hat{\nabla}_{\omega} L_{\nu_\rho^{\pi_\theta}}(\omega, \theta, \lambda)$ be the estimate produced by Algorithm \ref{algo_sampling}. Under assumptions \ref{ass_score} and \ref{ass_fisher}, the following semidefinite inequality holds for any $\theta\in\Theta$ and $\lambda\in \Lambda$.
\begin{align*}
    \mathbf{E}\left[\hat{\nabla}_{\omega} L_{\nu_\rho^{\pi_\theta}}(\omega^*_{\theta, \lambda}, \theta, \lambda) \otimes \hat{\nabla}_{\omega} L_{\nu_\rho^{\pi_\theta}}(\omega^*_{\theta, \lambda}, \theta, \lambda)\right] \preceq \sigma^2 F_\rho(\theta)
\end{align*}
where $\omega^*_{\theta, \lambda}$, $F_\rho(\theta)$ are given by \eqref{eq_quad_optimization} and \eqref{eq_def_F_rho} respectively and $\sigma^2$ is defined below.
\begin{align}
    \label{eq_def_sigma_2}
    \sigma^2\triangleq \dfrac{1}{(1-\gamma)^4}\left[\dfrac{2G^4}{\mu_F^2}+32\right] (1+\lambda_{\max})^2
\end{align}
\end{lemma}
The term $\sigma^2$ defined in Lemma \ref{lemma_5} can be described as the scaled variance of the gradient estimate, $\hat{\nabla}_{\omega}L_{\nu_\rho^{\pi_\theta}}(\omega, \theta, \lambda)$. Note that if the estimates were non-stochastic (i.e., deterministic), we would have $\sigma^2=0$ since $\nabla_\omega L_{\nu_\rho^{\pi_\theta}}(\omega_{\theta, \lambda}^*, \theta, \lambda)=0$. The last equation can be proved using the definition of $\omega_{\theta, \lambda}^*$ given in \eqref{eq_quad_optimization}, the gradient expression provided in \eqref{eq:quadratic_gradient}, and observing that the Fisher matrix, $F_\rho(\theta)$ is invertible due to Assumption \ref{ass_fisher}. The above information is crucial in bounding the first-order error, as stated in the following lemma.
\begin{lemma}
    \label{lemma_second_order} If assumptions \ref{ass_score} and \ref{ass_fisher} hold, then the following relations are satisfied $\forall k\in\{0,\cdots, K-1\}$ with learning rates $\alpha = \frac{3\sqrt{5}G^2}{\mu_F+3\sqrt{5}G^2}$, $\beta=\frac{\mu_F}{9G^2}$, $\xi=\frac{1}{3\sqrt{5}G^2}$, and $\delta=\frac{1}{5G^2}$ provided that the inner loop length of Algorithm \ref{algo_npg} obeys $H>\bar{C}\frac{G^2}{\mu_F}\log\left(\sqrt{\mathrm{d}}\frac{G^2}{\mu_F}\right)$ for some universal constant, $\bar{C}$.
    \begin{align}
    \label{eq:second_order_bound_lemma}
        &\mathbf{E}\Vert\omega_k - \omega_k^*\Vert^2 \leq 22\dfrac{\sigma^2\mathrm{d}}{\mu_F H}
        + C\exp\left(-\dfrac{\mu_F }{20G^2}H\right)\left[\dfrac{(1+\lambda_{\max})^2}{\mu_F(1-\gamma)^4}\right],\\
        \label{eq:first_order_bound_nested_exp_lemma}&\mathbf{E}\Vert\mathbf{E}\left[\omega_k\big|\theta_k, \lambda_k \right] - \omega_k^*\Vert \leq  \sqrt{C}\exp\left(-\dfrac{\mu_F }{40G^2}H\right)\left[\dfrac{1+\lambda_{\max}}{\sqrt{\mu_F}(1-\gamma)^2}\right]
    \end{align}
    where $C$ denotes a universal constant, $\omega_k$ is given by \eqref{eq:tail_average}, and $\sigma^2$ is defined in $\eqref{eq_def_sigma_2}$.
\end{lemma}

The first bound \eqref{eq:second_order_bound_lemma} is a consequence of Lemma \ref{lemma_5} and the ASGD convergence result provided in \citep{jain2018accelerating} (Corollary 2). To gain intuition about the second result \eqref{eq:first_order_bound_nested_exp_lemma}, note that by taking the conditional expectation $\mathbf{E}[\cdot|\theta_k, \lambda_k]$ on both sides of the ASGD iterations \eqref{eq:asgd_1}$-$\eqref{eq:asgd_4}, and applying the unbiasedness of the gradient estimate (Lemma \ref{lemma_unbiased}) we obtain the following $\forall h\in\{0, \cdots, H-1\}$.
\begin{align}
    \label{eq_asgd_det_1}
    \begin{split}
        & \bar{\mathbf{y}}_h = \alpha\bar{\mathbf{x}}_{h}+(1-\alpha)\bar{\mathbf{v}}_h, \\
        &\bar{\mathbf{x}}_{h+1}= \bar{\mathbf{y}}_h - \delta \nabla_{\omega}L_{\nu^{\pi_{\theta_k}}_\rho}(\omega, \theta_k, \lambda_k)\big|_{\omega = \bar{\mathbf{y}}_h}\\
        & \bar{\mathbf{z}}_h = \beta \bar{\mathbf{y}}_h + (1-\beta) \bar{\mathbf{v}}_h,\\  
        & \bar{\mathbf{v}}_{h+1}= \bar{\mathbf{z}}_h - \xi \nabla_{\omega}L_{\nu^{\pi_{\theta_k}}_\rho}(\omega, \theta_k, \lambda_k)\big|_{\omega=\bar{\mathbf{y}}_h}
    \end{split}
\end{align}
where $\bar{l}_h = \mathbf{E}[l_h|\theta_k, \lambda_k]$, $l\in\{\mathbf{v}, \mathbf{x}, \mathbf{y}, \mathbf{z}\}$. 
Moreover, taking conditional expectation on both sides of the tail averaging process \eqref{eq:tail_average}, we arrive at the following.
\begin{align}
    \bar{\omega}_k\triangleq \mathbf{E}\left[\omega_k\big|\theta_k, \lambda_k\right]= \dfrac{2}{H}\sum_{\frac{H}{2}<h\leq H} \bar{\mathbf{x}}_h
    \label{eq:tail_average_det}
\end{align}

Note that the steps \eqref{eq_asgd_det_1}$-$\eqref{eq:tail_average_det} resemble the iterative updates of a deterministic ASGD. This allows us to obtain $\mathbf{E}\Vert\bar{\omega}_k-\omega_k^*\Vert$ by substituting $\sigma^2=0$ in \eqref{eq:second_order_bound_lemma}
and applying the Cauchy-Schwarz inequality. 

\subsection{Global Convergence of the Lagrange}

Combining Lemma \ref{lemma_lagrange_convergence}, \eqref{eq_second_order_expand} and using the expected gradient errors provided by Lemma \ref{lemma_second_order}, we bound the average Lagrange optimality gap as a function of tunable parameters $H$ and $K$ as stated in the following corollary.

\begin{corollary}
    \label{corr_lagrange_global_convergence}
    Consider the same setup and the choice of parameters described in Lemma \ref{lemma_lagrange_convergence}$-$ \ref{lemma_second_order}. The following inequality holds if assumptions \ref{ass_score}$-$\ref{ass_fisher} are met.
    \begin{equation}
        \label{eq:general_bound_corr}
        \begin{split}
            &\frac{1}{K}\sum_{k=0}^{K-1}\mathbf{E}\bigg(J_{\mathrm{L}, \rho}(\pi^*, \lambda_k)-J_{\mathrm{L}, \rho}(\theta_k,\lambda_k)\bigg)\leq \sqrt{\epsilon_{\mathrm{bias}}} + \left[\dfrac{G\sqrt{C}(1+\lambda_{\max})}{\sqrt{\mu_F}(1-\gamma)^2}\right]\exp\left(-\dfrac{\mu_F }{40G^2}H\right)\\
            &+B\eta \left[\dfrac{22\sigma^2\mathrm{d}}{\mu_F H} + \exp\left(-\dfrac{\mu_F }{20G^2}H\right)\left[\dfrac{C(1+\lambda_{\max})^2}{\mu_F(1-\gamma)^4}\right]+ \dfrac{G^2(1+\lambda_{\max})^2}{\mu_F^2 (1-\gamma)^4}\right] \\
            & + \frac{1}{\eta K}\mathbf{E}_{s\sim d^{\pi^*}_\rho}[KL(\pi^*(\cdot\vert s)\Vert\pi_{\theta_0}(\cdot\vert s))]		
        \end{split}
    \end{equation}
\end{corollary}

Corollary \ref{corr_lagrange_global_convergence} bounds the optimality error of the Lagrange function as $\mathcal{O}(\sqrt{\epsilon_{\mathrm{bias}}}+\exp\left(-C_0H\right)+\eta+\frac{1}{\eta K})$ where $C_0$ is some problem specific constant. Interestingly, the dual learning parameter, $\zeta$ does not appear in this bound. However, the next section shows that $\zeta$ plays a pivotal role in deciding the objective and constraint violation rates.

\subsection{Decoupling the Objective and the Constraint Violation Rates}

The goal of the following theorem is to choose optimal values of the tunable parameters and decouple the objective and constraint violation rates from the Lagrange convergence result given in \eqref{eq:general_bound_corr}. 
\begin{theorem}
    \label{theorem_1}
    Consider the same setup and the choice of parameters given in Lemma \ref{lemma_lagrange_convergence}$-$ \ref{lemma_second_order}. Assume $\eta=(1-\gamma)^2(1+\lambda_{\max})^{-1}/\sqrt{K}$, $\zeta = \lambda_{\max}(1-\gamma)/\sqrt{K}$, and $\lambda_{\max}=2/[(1-\gamma)c_{\mathrm{slater}}]$. For sufficiently small $\epsilon>0$, the following inequalities hold
    \begin{align}
    \label{eq_regret_constraint_epsilon}
    \begin{split}
         &\frac{1}{K}\sum_{k=0}^{K-1}\mathbf{E}\bigg[J_{r, \rho}^{\pi^*}-J_{r, \rho}(\theta_k)\bigg]\leq \sqrt{\epsilon_{\mathrm{bias}}} + \epsilon, \\
         &\mathbf{E}\left[\dfrac{1}{K}\sum_{k=0}^{K-1}-J_{c, \rho}(\theta_k)\right] \leq (1-\gamma)c_{\mathrm{slater}} \sqrt{\epsilon_{\mathrm{bias}}} + \epsilon
    \end{split}
\end{align}
    whenever assumptions \ref{ass_slater}$-$\ref{ass_fisher} are met, $H=\mathcal{O}(\log(\epsilon^{-1}))$ and $K=\mathcal{O}((1-\gamma)^{-6}\epsilon^{-2})$. Therefore, the sample complexity to ensure \eqref{eq_regret_constraint_epsilon} is $\mathcal{O}((1-\gamma)^{-1}HK) = \tilde{\mathcal{O}}((1-\gamma)^{-7}\epsilon^{-2})$. It is to be clarified that the $(1-\gamma)^{-1}$ factor in the sample complexity calculation appears due to the fact that it requires $\mathcal{O}((1-\gamma)^{-1})$ samples on an average to obtain a gradient estimate via Algorithm \ref{algo_sampling}.
\end{theorem}

Theorem \ref{theorem_1} dictates that, with appropriate choice of the parameters, the rate of convergence of the objective and that of constraint violation can be bounded as $\mathcal{O}(\sqrt{\epsilon_{\mathrm{bias}}}+\epsilon)$ with $\tilde{\mathcal{O}}((1-\gamma)^{-7}\epsilon^{-2})$ samples. This beats the SOTA $\Tilde{\mathcal{O}}(\epsilon^{-4})$ sample complexity of \citep{bai2023achieving} and achieves the theoretical lower bound. Our derived sample complexity is also dependent on $c_{\mathrm{slater}}$. However, this is not explicitly mentioned in Theorem \ref{theorem_1}. Interested readers can find such details in the appendix.

\begin{remark}
    Note the importance of the nested expectation in the first-order term $\mathbf{E}\Vert\mathbf{E}[\omega_k|\theta_k, \lambda_k]-\omega_k^*\Vert$ in Lemma \ref{lemma_lagrange_convergence}. Lemma \ref{lemma_second_order} bounds this term as $\mathcal{O}(\exp(-C_0H))$ where $C_0$ is a problem dependent constant. Other terms in the Lagrange optimality bound (Lemma \ref{lemma_lagrange_convergence}) can be expressed as $\mathcal{O}(\eta+\frac{1}{\eta K})$. Moreover, following the proof of Theorem \ref{theorem_1}, one sees that decoupling the objective optimality error and constraint violation bounds incurs additional $\mathcal{O}(\zeta+\frac{1}{\zeta K})$ terms. Choosing $\eta, \zeta$ as prescribed in Theorem \ref{theorem_1}, makes both the objective and constraint violation errors as $\mathcal{O}(\sqrt{\epsilon_{\mathrm{bias}}}+\exp(-C_0 H)+K^{-0.5})$. This allows us to take $H=\tilde{\mathcal{O}}(1)$ and $K=\mathcal{O}(\epsilon^{-2})$, leading to $\tilde{\mathcal{O}}(\epsilon^{-2})$ sample complexity. 
    Had the first-order term been $\mathbf{E}\Vert\omega_k-\omega_k^*\Vert$, it would have resulted in $\mathcal{O}(\sqrt{\epsilon_{\mathrm{bias}}}+H^{-0.5}+K^{-0.5})$ objective and constraint violation errors which would have lead to $\mathcal{O}(\epsilon^{-4})$ sample complexity. 
\end{remark}

\section{Conclusions and Limitations}
\label{sec-conclusion}

This paper considers the problem of learning a CMDP where the goal is to maximize the objective value function while guaranteeing that the cost value exceeds a predefined threshold. We propose an acceleration-based primal-dual natural policy gradient algorithm that ensures $\epsilon$ optimality gap and $\epsilon$ constraint violation with $\tilde{\mathcal{O}}(\epsilon^{-2})$ sample complexity. This improves upon the previous state-of-the-art sample complexity of $\mathcal{O}(\epsilon^{-4})$ and achieves the theoretical lower bound. Future works include applying the idea of acceleration-based NPG to improve sample complexities in other related domains of constrained reinforcement learning, e.g., non-linear CMDP, average reward CMDP, etc.

\bibliographystyle{neurips} 
\bibliography{ref}

\newpage
\appendix

\section{Helper Lemma}
\begin{lemma}
    \label{lemma_helper}
    With slight abuse of notation, define $J_{\mathrm{L}, \rho}(\pi, \lambda) \triangleq J_{r, \rho}^{\pi}+\lambda J_{c, \rho}^{\pi}$. The following relation holds for any two policies $\pi_1, \pi_2$ and $\lambda\in \Lambda$.
    \begin{align}
        J_{\mathrm{L}, \rho}(\pi_1, \lambda) - J_{\mathrm{L}, \rho}(\pi_2, \lambda) = \dfrac{1}{1-\gamma} \mathbf{E}_{(s, a)\sim \nu^{\pi_1}_{\rho}}\left[ A^{\pi_2}_{\mathrm{L}, \lambda}(s, a)\right] 
    \end{align}
\end{lemma}
\begin{proof}
    This can be proved using Lemma 2 of \citep{agarwal2021theory} and the definition of the Lagrange function.
\end{proof}
\section{Proof of Lemma \ref{lemma_unbiased}}
\begin{proof}
    Fix arbitrary $\theta$ and $\lambda$. Note that the following equation holds due to the definition of $\hat{J}_{c, \rho}(\theta)$. 
    \begin{align}
        \begin{split}
            \mathbf{E}\left[\hat{J}_{c, \rho}(\theta)\big| \theta\right] &= (1-\gamma) \mathbf{E}\left[\sum_{t=0}^\infty \gamma^t \sum_{j=0}^{t} c(s_j, a_j)\bigg| s_0\sim\rho, \pi_\theta \right] \\
            &=(1-\gamma) \mathbf{E}\left[\sum_{j=0}^\infty  c(s_j, a_j)\sum_{t=j}^{\infty} \gamma^t \bigg| s_0\sim\rho, \pi_\theta \right]\\
            &=\mathbf{E}\left[\sum_{j=0}^\infty  \gamma^j c(s_j, a_j)  \bigg| s_0\sim\rho, \pi_\theta \right]= J_{c, \rho}(\theta)
        \end{split}
    \end{align}
    To prove the unbiasedness of the gradient, we first prove that the distribution of the sample pair $(\hat{s}, \hat{a})$ produced by Algorithm \ref{algo_sampling} is indeed $\nu_\rho^{\pi_\theta}$. Observe the following.
    \begin{align}
        \mathrm{Pr}(\hat{s}=s, \hat{a}=a|\rho, \pi_\theta) = (1-\gamma) \sum_{t=0}^\infty \gamma^t \mathrm{Pr}(s_t=s, a_t=a|s_0\sim \rho, \pi_\theta) = \nu_\rho^{\pi_\theta}(s, a)
    \end{align}
    Next, we show that for a given pair $(s, a)$, the estimate $\hat{Q}_g^{\pi_\theta}(s, a)$, $ g\in \{r, c\}$ is unbiased.
    \begin{align}
        \label{eq_appndx_unbiased_Q}
        \begin{split}
            \mathbf{E}\left[\hat{Q}_g^{\pi_\theta}(s, a)\big| \theta, s, a\right] &= (1-\gamma) \mathbf{E}\left[\sum_{t=0}^\infty \gamma^t \sum_{j=0}^t g(s_i, a_i)\bigg| s_0=s, a_0=a, \pi_\theta \right]\\
            &=(1-\gamma) \mathbf{E}\left[\sum_{j=0}^\infty  g(s_i, a_i) \sum_{t=j}^t \gamma^t\bigg| s_0=s, a_0=a, \pi_\theta \right]\\
            &= \mathbf{E}\left[\sum_{j=0}^\infty  \gamma^i g(s_i, a_i) \bigg| s_0=s, a_0=a, \pi_\theta \right] = Q^{\pi_\theta}_g(s, a)
        \end{split}
    \end{align}
    In a similar fashion, one can establish that $\mathbf{E}[\hat{V}^{\pi_\theta}_g(s)\big| \theta, s] = V^{\pi_\theta}_g(s)$, $\forall g\in\{r, c\}$. Combining this with \eqref{eq_appndx_unbiased_Q} leads to: $\mathbf{E}[\hat{A}^{\pi_\theta}_{\mathrm{L}, \lambda}(s, a)|\theta, \lambda, s, a] = A_{\mathrm{L}, \lambda}^{\pi_\theta}(s, a)$. We arrive at,
    \begin{align}
            \mathbf{E}\left[\hat{F}_\rho(\theta)\big|\theta\right] 
            &=\mathbf{E}_{(\hat{s}, \hat{a})\sim \nu_\rho^{\pi_\theta}} \left[\nabla_\theta \log \pi_\theta(\hat{a}|\hat{s})\otimes \nabla_\theta \log \pi_\theta (\hat{a}|\hat{s})\big| \theta\right] = F_\rho(\theta),\\
            \begin{split}
                \mathbf{E}\left[\hat{H}_\rho(\theta, \lambda)\big| \theta, \lambda\right] &= \mathbf{E}_{(\hat{s}, \hat{a})\sim \nu_\rho^{\pi_\theta}}\left[\mathbf{E}\left[\hat{A}^{\pi_\theta}_{\mathrm{L}, \lambda}(\hat{s}, \hat{a})\bigg|\theta, \lambda, \hat{s}, \hat{a}\right]\nabla_\theta \log \pi_\theta(\hat{a}|\hat{s})\bigg| \theta, \lambda \right]\\
                & =\mathbf{E}_{(\hat{s}, \hat{a})\sim \nu_\rho^{\pi_\theta}}\left[A^{\pi_\theta}_{\mathrm{L}, \lambda}(\hat{s}, \hat{a})\nabla_\theta \log \pi_\theta(\hat{a}|\hat{s})\big| \theta, \lambda \right] = H_\rho(\theta, \lambda)
            \end{split}
    \end{align}

    Finally, we arrive at the following by utilizing the definitions \eqref{eq:quadratic_gradient} and \eqref{eq:estimation_gradient}.
    \begin{align}
        \begin{split}
            \mathbf{E}\left[\hat{\nabla}_\omega L_{\nu_\rho^{\pi_\theta}}(\omega, \theta, \lambda)\big|\omega, \theta, \lambda\right] &= \mathbf{E}\left[\hat{F}_\rho(\theta)\big|\theta\right]\omega - \dfrac{1}{1-\gamma}\mathbf{E}\left[\hat{H}_\rho(\theta, \lambda)\big|\theta, \lambda\right]\\
            &=F_\rho(\theta)\omega - \dfrac{1}{1-\gamma}H_\rho(\theta, \lambda) = \nabla_\omega L_{\nu_\rho^{\pi_\theta}}(\omega, \theta, \lambda)
        \end{split}
    \end{align}
    This concludes the proof.
\end{proof}

\section{Proof of Lemma \ref{lemma_lagrange_convergence}}

\begin{proof}
    Invoking the definition of KL divergence, we arrive at the following series of inequalities.
    \begin{equation}
	\begin{aligned}
            &\mathbf{E}_{s\sim d^{\pi^*}_{\rho}}[KL(\pi^*(\cdot\vert s)\Vert\pi_{\theta_k}(\cdot\vert s))-KL(\pi^*(\cdot\vert s)\Vert\pi_{\theta_{k+1}}(\cdot\vert s))]\\
            &=\mathbf{E}_{(s, a)\sim \nu^{\pi^*}_{\rho}}\bigg[\log\frac{\pi_{\theta_{k+1}}(a\vert s)}{\pi_{\theta_k}(a\vert s)}\bigg]\\
            &\overset{(a)}\geq\mathbf{E}_{(s, a)\sim \nu^{\pi^*}_{\rho}}[\nabla_\theta\log\pi_{\theta_k}(a\vert s)\cdot(\theta_{k+1}-\theta_k)]-\frac{B}{2}\Vert\theta_{k+1}-\theta_k\Vert^2\\
            &=\eta\mathbf{E}_{(s, a)\sim \nu^{\pi^*}_{\rho}}[\nabla_{\theta}\log\pi_{\theta_k}(a\vert s)\cdot\omega_k]-\frac{B\eta^2}{2}\Vert\omega_k\Vert^2\\
            &=\eta\mathbf{E}_{(s, a)\sim \nu^{\pi^*}_{\rho}}[\nabla_\theta\log\pi_{\theta_k}(a\vert s)\cdot\omega^*_k] + \eta\mathbf{E}_{(s, a)\sim \nu^{\pi^*}_{\rho}}[\nabla_\theta\log\pi_{\theta_k}(a\vert s)\cdot(\omega_k-\omega^*_k)]-\frac{B\eta^2}{2}\Vert\omega_k\Vert^2\\
            &=\eta[J_{\mathrm{L}, \rho}(\pi^*,\lambda_k)-J_{\mathrm{L}, \rho}(\theta_k,\lambda_k)]+\eta\mathbf{E}_{(s, a)\sim \nu^{\pi^*}_{\rho}}[\nabla_\theta\log\pi_{\theta_k}(a\vert s)\cdot\omega^*_k]\\
            &-\eta[J_{\mathrm{L}, \rho}(\pi^*,\lambda_k)-J_{\mathrm{L}, \rho}(\theta_k,\lambda_k)]
            +\eta\mathbf{E}_{(s, a)\sim \nu^{\pi^*}}[\nabla_\theta\log\pi_{\theta_k}(a\vert s)\cdot(\omega_k-\omega^*_k)]-\frac{B\eta^2}{2}\Vert\omega_k\Vert^2\\		
            &\overset{(b)}=\eta[J_{\mathrm{L}, \rho}(\pi^*,\lambda_k)-J_{\mathrm{L}, \rho}(\theta_k,\lambda_k)]+\eta\mathbf{E}_{(s, a)\sim \nu^{\pi^*}_{\rho}}\bigg[\nabla_\theta\log\pi_{\theta_k}(a\vert s)\cdot\omega^*_k-\dfrac{1}{1-\gamma}A_{\mathrm{L},\lambda_k}^{\pi_{\theta_k}}(s,a)\bigg]\\
            &+\eta\mathbf{E}_{(s, a)\sim \nu^{\pi^*}_{\rho}}[\nabla_\theta\log\pi_{\theta_k}(a\vert s)\cdot(\omega_k-\omega^*_k)]-\frac{B\eta^2}{2}\Vert\omega_k\Vert^2\\
            &\overset{(c)}\geq\eta[J_{\mathrm{L}, \rho}(\pi^*,\lambda_k)-J_{\mathrm{L}, \rho}(\theta_k,\lambda_k)]-\eta\sqrt{\mathbf{E}_{(s, a)\sim \nu^{\pi^*}_{\rho}}\bigg[\nabla_\theta\log\pi_{\theta_k}(a\vert s)\cdot\omega^*_k-\dfrac{1}{1-\gamma}A_{\mathrm{L},\lambda_k}^{\pi_{\theta_k}}(s,a)\bigg]^2}\\
            &+\eta\mathbf{E}_{(s, a)\sim \nu^{\pi^*}_{\rho}}[\nabla_\theta\log\pi_{\theta_k}(a\vert s)\cdot(\omega_k-\omega^*_k)]-\frac{B\eta^2}{2}\Vert\omega_k\Vert^2\\
            &\overset{(d)}\geq\eta[J_{\mathrm{L}, \rho}(\pi^*,\lambda_k)-J_{\mathrm{L}, \rho}(\theta_k,\lambda_k)]-\eta\sqrt{\epsilon_{\mathrm{bias}}}\\
            &\hspace{3cm}+\eta\mathbf{E}_{(s, a)\sim \nu^{\pi^*}_{\rho}}[\nabla_\theta\log\pi_{\theta_k}(a\vert s)\cdot(\omega_k-\omega^*_k)]-\frac{B\eta^2}{2}\Vert\omega_k\Vert^2\\
	\end{aligned}	
    \end{equation}
    where the step (a) holds by Assumption \ref{ass_score} and step (b) holds by Lemma \ref{lemma_helper}. Step (c) uses the convexity of the function $f(x)=x^2$. Finally, step (d) comes from the Assumption \ref{ass_transfer_error}. Rearranging the terms and taking expectations on both sides, we have,
    \begin{equation}
	\begin{split}
            &\mathbf{E}\left[ J_{\mathrm{L}, \rho}(\pi^*,\lambda_k)-J_{\mathrm{L}, \rho}(\theta_k,\lambda_k)\right] \leq  - \mathbf{E}_{(s, a)\sim \nu^{\pi^*}_{\rho}}\mathbf{E}\left[\nabla_\theta\log\pi_{\theta_k}(a\vert s)\cdot(\mathbf{E}\left[\omega_k\big|\theta_k, \lambda_k\right]-\omega^*_k)\right]\\
            &+\frac{B\eta}{2}\mathbf{E}\Vert\omega_k\Vert^2
            +\frac{1}{\eta}\mathbf{E}_{s\sim d^{\pi^*}_{\rho}}\left[\mathbf{E}\left[KL(\pi^*(\cdot\vert s)\Vert\pi_{\theta_k}(\cdot\vert s))\right]-\mathbf{E}\left[KL(\pi^*(\cdot|s)\Vert\pi_{\theta_{k+1}}(\cdot\vert s))\right]\right]+\sqrt{\epsilon_{\mathrm{bias}}}\\
            & \overset{(a)}{\leq} \sqrt{\epsilon_{\mathrm{bias}}} +  G\mathbf{E}\left\Vert(\mathbf{E}\left[\omega_k\big|\theta_k, \lambda_k\right]-\omega^*_k)\right\Vert + \frac{B\eta}{2}\mathbf{E}\Vert\omega_k\Vert^2 \\
            & +\frac{1}{\eta}\mathbf{E}_{s\sim d^{\pi^*}_{\rho}}\left[\mathbf{E}\left[KL(\pi^*(\cdot\vert s)\Vert\pi_{\theta_k}(\cdot\vert s))\right]-\mathbf{E}\left[KL(\pi^*(\cdot|s)\Vert\pi_{\theta_{k+1}}(\cdot\vert s))\right]\right]
	\end{split}
    \end{equation}
    where (a) follows from Assumption \ref{ass_score}. Summing from $k=0$ to $K-1$, using the non-negativity of KL divergence and dividing the resulting expression by $K$, we obtain,
    \begin{equation}
        \label{eq:general_bound_appndx_repeat}
        \begin{split}
            \frac{1}{K}\sum_{k=0}^{K-1}\mathbf{E}\bigg(J_{\mathrm{L}, \rho}(\pi^*, \lambda_k)&-J_{\mathrm{L}, \rho}(\theta_k,\lambda_k)\bigg)\leq \sqrt{\epsilon_{\mathrm{bias}}}+\frac{G}{K}\sum_{k=0}^{K-1}\mathbf{E}\Vert(\mathbf{E}\left[\omega_k\big|\theta_k, \lambda_k\right]-\omega^*_k)\Vert\\
            &+\frac{B\eta}{2K}\sum_{k=0}^{K-1}\mathbf{E}\Vert\omega_k\Vert^2
            +\frac{1}{\eta K}\mathbf{E}_{s\sim d^{\pi^*}_\rho}[KL(\pi^*(\cdot\vert s)\Vert\pi_{\theta_0}(\cdot\vert s))]		
        \end{split}
    \end{equation}
    This concludes the proof.
\end{proof}
\section{Proof of Lemma \ref{lemma_5}}

\begin{proof}
    Fix a $\theta\in\Theta$ and a $\lambda\in\Lambda$. Observe the following equation.
    \begin{align*}
        \begin{split}
            &\mathbf{E}\left[\hat{\nabla}_{\omega}L_{\nu^{\pi_\theta}_\rho}(\omega_{\theta, \lambda}^*, \theta, \lambda)\otimes\hat{\nabla}_{\omega}L_{\nu^{\pi_\theta}_\rho}(\omega_{\theta, \lambda}^*, \theta, \lambda)\right] \\
            &= \mathbf{E}_{(s, a)\sim \nu_\rho^{\pi_{\theta}}}\bigg[\mathbf{E}\bigg[\underbrace{\nabla_\theta\log\pi_{\theta}(a\vert s)\cdot\omega^*_{\theta, \lambda}-\dfrac{1}{1-\gamma}\hat{A}_{\mathrm{L}, \lambda}^{\pi_{\theta}}(s,a)}_{\triangleq \zeta_{\theta, \lambda}(s, a)}\bigg]^2\nabla_\theta\log\pi_\theta(a|s)\otimes \nabla_\theta\log\pi_\theta(a|s)\bigg]
        \end{split}
    \end{align*}

    To prove the lemma, it is sufficient to demonstrate that $\mathbf{E}[\zeta_{\theta, \lambda}(s, a)]\leq \sigma^2$, $\forall (s, a)$. Notice the chain of inequalities stated below.
    \begin{align}
        \label{eq:eq_41_appndx}
        \begin{split}
            \mathbf{E}&\bigg[\nabla_\theta\log\pi_{\theta}(a\vert s)\cdot\omega^*_{\theta, \lambda}-\dfrac{1}{1-\gamma}\hat{A}_{\mathrm{L}, \lambda}^{\pi_{\theta}}(s,a)\bigg]^2\\
            &\leq 2\left[\nabla_\theta\log\pi_{\theta}(a\vert s)\cdot\omega^*_{\theta, \lambda}\right]^2 + \dfrac{2}{(1-\gamma)^2} \mathbf{E}\left[\hat{A}^{\pi_\theta}_{\mathrm{L}, \lambda}(s, a)\right]^2\\
            &\overset{(a)}{\leq} 2\Vert\nabla_\theta\log\pi_\theta(a|s)\Vert^2 \Vert\omega_{\theta, \lambda}^*\Vert^2 + \dfrac{4(1+\lambda)^2}{(1-\gamma)^2} \max_{g\in\{r, c\}}\left\{\mathbf{E}\left[\hat{A}^{\pi_\theta}_{g}(s, a)\right]^2\right\} \\
            &\overset{(b)}{\leq} 2G^2\Vert F_\rho(\theta)^{\dagger}\nabla_\theta J_{\mathrm{L},\rho}(\theta, \lambda)\Vert^2 + \dfrac{8(1+\lambda)^2}{(1-\gamma)^2}\max_{g\in\{r, c\}}\left\{\mathbf{E}\left[\hat{Q}^{\pi_\theta}_{g}(s, a)\right]^2+\mathbf{E}\left[\hat{V}^{\pi_\theta}_{g}(s, a)\right]^2\right\} \\
            &\overset{(c)}{\leq} \dfrac{2G^4(1+\lambda)^2}{\mu_F^2(1-\gamma)^4} + \dfrac{32(1+\lambda)^2}{(1-\gamma)^4}\leq \dfrac{1}{(1-\gamma)^4}\left[\dfrac{2G^4}{\mu_F^2}+32\right] (1+\lambda_{\max})^2
        \end{split}
    \end{align}
    Inequality $(a)$ follows from the Cauchy-Schwarz inequality and the fact that $(a+b)^2\leq 2(a^2+b^2)$ for any two numbers $a, b$. The same argument is also applied in inequality $(b)$. Additionally, it uses Assumption \ref{ass_score} and the definition of $\omega_{\theta, \lambda}^*$.  Finally, $(c)$ is a consequence of Assumption \ref{ass_fisher}, Lemma \ref{lemma_2}, and the following two bounds.
\begin{align}
\label{eq:eq_42}
    \mathbf{E}\left[\hat{Q}_g^{\pi_\theta}(s, a)\right]^2\leq \dfrac{2}{(1-\gamma)^2}, ~\text{and}~\mathbf{E}\left[\hat{V}_g^{\pi_\theta}(s)\right]^2\leq \dfrac{2}{(1-\gamma)^2}, ~~\forall(s, a)\in \mathcal{S}\times\mathcal{A}, \forall g\in \{r, c\}
\end{align}
To establish the first bound, note that $|\hat{Q}_g^{\pi_{\theta}}(s, a)|$ is assigned a value of at most $(j+1)$ with probability $(1-\gamma)\gamma^j$, $\forall g\in\{r,c\}$. Therefore,
\begin{align}
    \mathbf{E}\left[\hat{Q}_g^{\pi_\theta}(s, a)\right]^2\leq \sum_{j=0}^\infty (1-\gamma)(j+1)^2 \gamma^j = \dfrac{(1+\gamma)}{(1-\gamma)^2} < \dfrac{2}{(1-\gamma)^2}
\end{align}
The second bound in $(\ref{eq:eq_42})$ can be proven similarly. This concludes the lemma.
\end{proof}
\section{Proof of Lemma \ref{lemma_second_order}}

We establish Lemma \ref{lemma_second_order} applying Corollary 2 of \citep{jain2018accelerating}. Note the following statements.

$\mathbf{S1}:$ The following quantities exist and are finite $\forall \theta\in\Theta$.
\begin{align}
    &F_{\rho}(\theta) \triangleq \mathbf{E}_{(s, a)\sim \nu^{\pi_{\theta}}_\rho}\big[\nabla_\theta \log\pi_{\theta}(a|s)\otimes \nabla_\theta \log\pi_{\theta}(a|s)\big],\\
    &G_{\rho}(\theta) \triangleq \mathbf{E}_{(s, a)\sim \nu^{\pi_{\theta}}_\rho}\big[\nabla_\theta \log\pi_{\theta}(a|s)\otimes \nabla_\theta \log\pi_{\theta}(a|s)\otimes \nabla_\theta \log\pi_{\theta}(a|s)\otimes \nabla_\theta \log\pi_{\theta}(a|s)\big]
\end{align}
$\mathbf{S2}:$ There exists $\sigma^2$ such that the following is obeyed $\forall \theta \in \Theta$ where $\omega_{\theta, \lambda}^*$ minimizes  $L_{\nu^{\pi_\theta}_\rho}(\cdot, \theta, \lambda)$.
\begin{align}
    \label{eq:eq_46_appndx}
    \begin{split}
        \mathbf{E}&\left[\hat{\nabla}_{\omega}L_{\nu^{\pi_\theta}_\rho}(\omega_{\theta, \lambda}^*, \theta, \lambda)\otimes\hat{\nabla}_{\omega}L_{\nu^{\pi_\theta}_\rho}(\omega_{\theta, \lambda}^*, \theta, \lambda)\right]\preccurlyeq \sigma^2 F_\rho(\theta)
    \end{split}
\end{align}
$\mathbf{S3:}$ There exists $\mu_F, G>0$ such that the following statements hold $\forall \theta \in \Theta$.
\begin{align}
    (a)~&F_\rho(\theta)\succcurlyeq \mu_F I_{\mathrm{d}},\\
    (b)~& \mathbf{E}_{(s, a)\sim \nu^{\pi_\theta}_\rho}\left[\Vert \nabla_\theta \log\pi_{\theta}(a|s) \Vert^2 \nabla_\theta \log\pi_{\theta}(a|s)\otimes \nabla_\theta \log\pi_{\theta}(a|s)\right]\preccurlyeq G^2 F_\rho(\theta), \\
    (c)~& \mathbf{E}_{(s, a)\sim \nu^{\pi_\theta}_\rho}\left[\Vert \nabla_\theta \log\pi_{\theta}(a|s) \Vert^2_{F_\rho(\theta)^{\dagger}} \nabla_\theta \log\pi_{\theta}(a|s)\otimes \nabla_\theta \log\pi_{\theta}(a|s)\right]\preccurlyeq \dfrac{G^2}{\mu_F} F_\rho(\theta)
\end{align}
Statement $\mathbf{S1}$ follows from Assumption \ref{ass_score} whereas $\mathbf{S2}$ is a consequence of Lemma \ref{lemma_5}. Statement $\mathbf{S3}(a)$ is identical to Assumption \ref{ass_fisher}, $\mathbf{S3}(b)$ results from Assumption \ref{ass_score}, and finally, $\mathbf{S3}(c)$ follows from Assumption \ref{ass_score} and \ref{ass_fisher}. We can, therefore, apply Corollary 2 of \citep{jain2018accelerating} with $\kappa=\Tilde{\kappa}=G^2/\mu_F$ and deduce the following convergence result whenever $H>\bar{C}\sqrt{\kappa\Tilde{\kappa}}\log(\sqrt{\mathrm{d}}\sqrt{\kappa\Tilde{\kappa}})$  and the learning rates are set as $\alpha = \frac{3\sqrt{5} \sqrt{\kappa\Tilde{\kappa}}}{1+3\sqrt{5\kappa\Tilde{\kappa}}}$, $\beta = \frac{1}{9\sqrt{\kappa\Tilde{\kappa}}}$, $\xi=\frac{1}{3\sqrt{5}\mu_F\sqrt{\kappa\Tilde{\kappa}}}$, and $\delta = \frac{1}{5G^2}$. 
\begin{align}
\label{eq:eq_50_appndx}
\begin{split}
    & \mathbf{E}\left[l_k(\omega_k)\right] - l_k(\omega_k^*) \leq \dfrac{C}{2}\exp\left(-\dfrac{H}{20\sqrt{\kappa\Tilde{\kappa}}}\right)\left[l_k(\mathbf{0})-l_k(\omega_k^*)\right]+11\dfrac{\sigma^2\mathrm{d}}{H}, \\
    &\text{where~} l_k(\omega) \triangleq L_{\nu^{\pi_{\theta_k}}_\rho}(\omega, \theta_k, \lambda_k), ~\forall \omega\in\mathbb{R}^{\mathrm{d}}
\end{split}
\end{align}
The term, $C$ is a universal constant. Note that $l_k(\omega_k^*)\geq 0$ and $l_k(\mathbf{0})$ is bounded above as follows. 
\begin{align}
\label{eq:eq_51_appndx}
    \begin{split}
        l_k(\mathbf{0})=\dfrac{1}{2}\mathbf{E}_{(s, a)\sim \nu^{\pi_{\theta_k}}_\rho}&\bigg[\dfrac{1}{1-\gamma}A_{\mathrm{L}, \lambda_k}^{\pi_{\theta_k}}(s,a)\bigg]^2\overset{(a)}{\leq} \dfrac{(1+\lambda_{\max})^2}{2(1-\gamma)^4}
    \end{split}
\end{align}
where $(a)$ is a result of the fact that $|A_{\mathrm{L}, \lambda}^{\pi_{\theta}}(s, a)|\leq (1+\lambda_{\max})/(1-\gamma)$, $\forall (s, a)\in \mathcal{S}\times \mathcal{A}$, $\forall \theta\in \Theta$, and $\forall \lambda\in \Lambda$. Combining $\eqref{eq:eq_50_appndx}$,  $\eqref{eq:eq_51_appndx}$, and the fact that $l_k(\cdot)$ is $\mu_F$-strongly convex, we establish,
    \begin{align}
    \label{eq:eq_appndx_52}
    \begin{split}
        \mathbf{E}\Vert\omega_k - \omega_k^*\Vert^2 \leq \frac{2}{\mu_F} \big[\mathbf{E}\left[l_k(\omega_k)\right] - l_k(\omega_k^*)\big]&\leq 22\dfrac{\sigma^2\mathrm{d}}{\mu_F H}
        + C\exp\left(-\dfrac{\mu_F }{20G^2}H\right)\left[\dfrac{(1+\lambda_{\max})^2}{\mu_F(1-\gamma)^4}\right]
    \end{split}
    \end{align}
This proves the first statement. We get the following for noiseless ($\sigma^2=0$) gradient updates.
    \begin{align}
    \label{eq:eq_appndx_53}
    \begin{split}
        \mathbf{E}\Vert(\mathbf{E}[\omega_k|\theta_k] - \omega_k^*)\Vert^2 &\leq  C\exp\left(-\dfrac{\mu_F }{20G^2}H\right)\left[\dfrac{(1+\lambda_{\max})^2}{\mu_F(1-\gamma)^4}\right]
    \end{split}
    \end{align}
The second statement can be established from $\eqref{eq:eq_appndx_53}$ by applying Jensen's inequality on the function $f(x)=x^2$.

\section{Proof of Theorem \ref{theorem_1}}

Applying the inequalities $H\geq 1$, $\exp(-(\mu_F/20G^2)H)\leq 1$ and substituting the values of $\sigma^2$ and $\eta$ (as stated in Lemma \ref{lemma_5} and Theorem \ref{theorem_1} respectively), we can rewrite \eqref{eq:general_bound_corr} as follows.
\begin{align}
    \label{eq_appndx_57}
    \begin{split}
        \frac{1}{K}\sum_{k=0}^{K-1}\mathbf{E}\bigg(J_{\mathrm{L}, \rho}(\pi^*, \lambda_k)&-J_{\mathrm{L}, \rho}(\theta_k,\lambda_k)\bigg)\\
        &\leq \sqrt{\epsilon_{\mathrm{bias}}} + f_0\dfrac{(1+\lambda_{\max})}{(1-\gamma)^2}\exp\left(-\dfrac{\mu_F}{40G^2}H\right) + f_1\dfrac{(1+\lambda_{\max})}{(1-\gamma)^2}\dfrac{1}{\sqrt{K}} 
    \end{split}
\end{align}

The terms $f_0$ and $f_1$ are defined below.
\begin{align}
    \begin{split}
        & f_0\triangleq \dfrac{G\sqrt{C}}{\sqrt{\mu_F}},\\
        & f_1 \triangleq B\left[\dfrac{44\mathrm{d}}{\mu_F}\left(\dfrac{G^4}{\mu_F^2}+16\right)+\dfrac{C}{\mu_F}+\dfrac{G^2}{\mu_F^2}\right]+ \mathbf{E}_{s\sim d^{\pi^*}_\rho}[KL(\pi^*(\cdot\vert s)\Vert\pi_{\theta_0}(\cdot\vert s))]
    \end{split}
\end{align}

Using the definition of $J_{\mathrm{L}, \rho}(\cdot, \cdot)$, one can write,
\begin{align}
    \label{eq_appndx_59}
    \begin{split}
        J_{\mathrm{L}, \rho}(\pi^*, \lambda_k)-J_{\mathrm{L}, \rho}(\theta_k,\lambda_k) &= (J_{r, \rho}^{\pi^*} - J_{r, \rho}(\theta_k))+\lambda_k(J_{c, \rho}^{\pi^*} - J_{c, \rho}(\theta_k))\\
        &\overset{(a)}{\geq} (J_{r, \rho}^{\pi^*} - J_{r, \rho}(\theta_k))+\lambda_k( - J_{c, \rho}(\theta_k))
    \end{split}
\end{align}
where (a) follows from the fact that $\lambda_k\geq 0$ and $J_{c, \rho}^{\pi^*}\geq 0$ due to feasibility. Combining \eqref{eq_appndx_57} and \eqref{eq_appndx_59}, we obtain the following.
\begin{align}
    \label{eq_appndx_60}
    \begin{split}
        \frac{1}{K}\sum_{k=0}^{K-1}\mathbf{E}\bigg[J_{r, \rho}^{\pi^*}&-J_{r, \rho}(\theta_k)\bigg]+\dfrac{1}{K}\sum_{k=0}^{K-1}\mathbf{E}\bigg[-\lambda_k J_{c, \rho}(\theta_k)\bigg]\\
        &\leq \sqrt{\epsilon_{\mathrm{bias}}} + f_0\dfrac{(1+\lambda_{\max})}{(1-\gamma)^2}\exp\left(-\dfrac{\mu_F}{40G^2}H\right) + f_1\dfrac{(1+\lambda_{\max})}{(1-\gamma)^2}\dfrac{1}{\sqrt{K}} 
    \end{split}
\end{align}

\subsection{Convergence Rate of the Objective Function}

Note that \eqref{eq_appndx_60} can be alternatively written as,
\begin{align}
    \label{eq_appndx_62}
    \begin{split}
        \frac{1}{K}\sum_{k=0}^{K-1}\mathbf{E}\bigg[J_{r, \rho}^{\pi^*}&-J_{r, \rho}(\theta_k)\bigg]\leq \sqrt{\epsilon_{\mathrm{bias}}} + f_0\dfrac{(1+\lambda_{\max})}{(1-\gamma)^2}\exp\left(-\dfrac{\mu_F}{40G^2}H\right) \\
        &+ f_1\dfrac{(1+\lambda_{\max})}{(1-\gamma)^2}\dfrac{1}{\sqrt{K}} + \dfrac{1}{K}\sum_{k=0}^{K-1}\mathbf{E}\bigg[\lambda_k J_{c, \rho}(\theta_k)\bigg]
    \end{split}
\end{align}

To obtain the convergence rate of the objective function, we need to bound the last term in the above expression. Observe the following chain of inequalities.
\begin{equation}
    \label{eq:bound_lambdak_appndx}
    \begin{aligned}
        0\leq &(\lambda_{K})^2\overset{(a)}{=}\sum_{k=0}^{K-1}\bigg((\lambda_{k+1})^2-(\lambda_{k})^2\bigg)\\
        &\overset{(b)}{\leq}\sum_{k=0}^{K-1}\bigg(\big[\lambda_{k}-\zeta\hat{J}_{c, \rho}(\theta_k)\big]^2-(\lambda_{k})^2\bigg)
        =-2\zeta\sum_{k=0}^{K-1}\lambda_{k}\hat{J}_{c, \rho}(\theta_k)+\zeta^2\sum_{k=0}^{K-1}\hat{J}^2_{c, \rho}(\theta_k)
    \end{aligned}
\end{equation}
where (a) uses $\lambda_0=0$ and (b) follows from the contraction property of the projection operator, $\mathcal{P}_{\Lambda}$. Using the above inequality, one can write,
\begin{align}
    \label{eq_appndx_64}
    \dfrac{1}{K}\sum_{k=0}^{K-1} \mathbf{E}\bigg[\lambda_k \hat{J}_{c, \rho}(\theta_k)\bigg] \leq \dfrac{\zeta}{2K}\sum_{k=0}^{K-1} \mathbf{E}\bigg[\hat{J}^2_{c, \rho}(\theta_k)\bigg]
\end{align}

Using the unbiasedness of $\hat{J}_{c, \rho}(\theta_k)$ (Lemma \ref{lemma_unbiased}), we deduce the following.
\begin{align}
    \label{eq_appndx_65}
    \mathbf{E}\bigg[\lambda_k\hat{J}_{c, \rho}(\theta_k)\bigg] \overset{(a)}{=} \mathbf{E}\bigg[\lambda_k\mathbf{E}\left[\hat{J}_{c, \rho}(\theta_k)\big| \theta_k\right]\bigg] =  \mathbf{E}\bigg[\lambda_kJ_{c, \rho}(\theta_k)\bigg]
\end{align}
where $(a)$ is a consequence of the fact that $\hat{J}_{c, \rho}(\theta_k)$ and $\lambda_k$ are conditionally independent given $\theta_k$. Note that, $\hat{J}^2_{c, \rho}(\theta_k)$ is assigned a value of at most $(j+1)^2$ with probability $(1-\gamma)\gamma^j$, $j\in\{0, 1, \cdots\}$. Therefore, the RHS of \eqref{eq_appndx_64} can be bounded as follows.
\begin{align}
    \label{eq_appndx_66}
    \mathbf{E}\bigg[\hat{J}^2_{c, \rho}(\theta_k)\bigg] \leq \sum_{j=0}^{\infty}(1-\gamma)(j+1)^2 \gamma^j = \dfrac{1+\gamma}{(1-\gamma)^2} < \dfrac{2}{(1-\gamma)^2}
\end{align}

Combining \eqref{eq_appndx_62}, \eqref{eq_appndx_64}, \eqref{eq_appndx_65}, and \eqref{eq_appndx_66}, we finally obtain,
\begin{align}
    \label{eq_appndx_67}
    \begin{split}
        &\frac{1}{K}\sum_{k=0}^{K-1}\mathbf{E}\bigg[J_{r, \rho}^{\pi^*}-J_{r, \rho}(\theta_k)\bigg]\\
        &\leq \sqrt{\epsilon_{\mathrm{bias}}} + f_0\dfrac{(1+\lambda_{\max})}{(1-\gamma)^2}\exp\left(-\dfrac{\mu_F}{40G^2}H\right)
        + f_1\dfrac{(1+\lambda_{\max})}{(1-\gamma)^2}\dfrac{1}{\sqrt{K}} + \dfrac{\zeta}{(1-\gamma)^2} \\
        &\overset{(a)}{=} \sqrt{\epsilon_{\mathrm{bias}}} + f_0\dfrac{(1+\lambda_{\max})}{(1-\gamma)^2}\exp\left(-\dfrac{\mu_F}{40G^2}H\right)
        + f_1\dfrac{(1+\lambda_{\max})}{(1-\gamma)^2}\dfrac{1}{\sqrt{K}} + \dfrac{\lambda_{\max}}{(1-\gamma)} \dfrac{1}{\sqrt{K}}
        \\
        &\leq \sqrt{\epsilon_{\mathrm{bias}}} + f_0\dfrac{(1+\lambda_{\max})}{(1-\gamma)^2}\exp\left(-\dfrac{\mu_F}{40G^2}H\right)
        + (f_1+1)\dfrac{(1+\lambda_{\max})}{(1-\gamma)^2}\dfrac{1}{\sqrt{K}} 
    \end{split}
\end{align}
where $(a)$ uses the substitution $\zeta = \lambda_{\max}(1-\gamma)/\sqrt{K}$.

\subsection{Rate of Constraint Violation}

The following inequality is satisfied for any $k\in\{0, 1, \cdots, K-1\}$.
\begin{equation}
    \begin{aligned}
	    &\vert\lambda_{k+1} - \lambda_{\max}\vert^2=|\mathcal{P}_{\Lambda}(\lambda_k-\zeta \hat{J}_{c, \rho}(\theta_k))-\lambda_{\max}|^2\\ 
		&\overset{(a)}{\leq} \big|\lambda_{k} - \zeta \hat{J}_{c, \rho}(\theta_{k})   -\lambda_{\max}\big|^2
		=\big|\lambda_{k} -\lambda_{\max}\big|^2 -2\zeta \hat{J}_{c, \rho}(\theta_k)\big(\lambda_{k}  -\lambda_{\max}\big) +\zeta^2 \hat{J}^2_{c, \rho}(\theta_{k})
	\end{aligned}
\end{equation}
where (a) is due to the contractive property of $\mathcal{P}_{\Lambda}$. Performing an average over $k\in\{0, \cdots, K-1\}$, and applying expectations on both sides, we get,
\begin{align}
    \label{eq_appndx_69}
    \begin{split}
        \dfrac{1}{K}\sum_{k=0}^{K-1}\mathbf{E}\left[(\lambda_k-\lambda_{\max})\hat{J}_{c, \rho}(\theta_k)\right]&\overset{(a)}{\leq} \dfrac{|\lambda_{\max}|^2-|\lambda_{K}-\lambda_{\max}|^2}{2\zeta K} + \dfrac{\zeta}{2K}\sum_{k=0}^{K-1}\mathbf{E}\left[\hat{J}^2_{c, \rho}(\theta_k)\right]\\
        &\overset{(b)}{\leq} \dfrac{\lambda_{\max}^2}{2\zeta K}+\dfrac{\zeta}{(1-\gamma)^2}\overset{(c)}{=} \dfrac{3\lambda_{\max}}{2(1-\gamma)}\dfrac{1}{\sqrt{K}}
    \end{split}
\end{align}
where $(a)$ utilises $\lambda_0=0$, $(b)$ applies \eqref{eq_appndx_66}, and $(c)$ is derived using $\zeta = \lambda_{\max}(1-\gamma)/\sqrt{K}$. Note that one can write the following using Lemma \ref{lemma_unbiased} and the observation that $\hat{J}_{c, \rho}(\theta_k)$ and $\lambda_k$ are conditionally independent given $\theta_k$.
\begin{align}
    \label{eq_appndx_70}
    \begin{split}
        \mathbf{E}\left[(\lambda_k-\lambda_{\max})\hat{J}_{c, \rho}(\theta_k)\right]
        =\mathbf{E}\left[(\lambda_k-\lambda_{\max})\mathbf{E}\left[\hat{J}_{c, \rho}(\theta_k)\big|\theta_k\right]\right]
        \overset{}{=} \mathbf{E}\bigg[(\lambda_k-\lambda_{\max})J_{c, \rho}(\theta_k)\bigg]
    \end{split}
\end{align}
Combining \eqref{eq_appndx_69} and \eqref{eq_appndx_70}, we get,
\begin{align}
    \label{eq_appndx_71}
    \dfrac{1}{K}\sum_{k=0}^{K-1} \mathbf{E}\bigg[(\lambda_k-\lambda_{\max})J_{c, \rho}(\theta_k)\bigg] \leq \dfrac{3\lambda_{\max}}{2(1-\gamma)}\dfrac{1}{\sqrt{K}}\leq \dfrac{2(1+\lambda_{\max})}{(1-\gamma)^2 \sqrt{K}}
\end{align}
Finally, combining \eqref{eq_appndx_62} and \eqref{eq_appndx_71}, we arrive at,
\begin{align}
    \label{eq_appndx_72}
    \begin{split}
        \frac{1}{K}\sum_{k=0}^{K-1}\mathbf{E}\bigg[J_{r, \rho}^{\pi^*}&-J_{r, \rho}(\theta_k)\bigg]+\lambda_{\max}\dfrac{1}{K}\sum_{k=0}^{K-1}\mathbf{E}\bigg[- J_{c, \rho}(\theta_k)\bigg]\\
        &\leq \sqrt{\epsilon_{\mathrm{bias}}} + f_0\dfrac{(1+\lambda_{\max})}{(1-\gamma)^2}\exp\left(-\dfrac{\mu_F}{40G^2}H\right) + (f_1+2)\dfrac{(1+\lambda_{\max})}{(1-\gamma)^2}\dfrac{1}{\sqrt{K}} 
    \end{split}
\end{align}

Since the functions $\{J_{g, \rho}(\theta_k)\}$, $g\in\{r, c\}$, $k\in\{0, \cdots, K-1\}$ are linear in occupancy measure, there exists a policy $\bar{\pi}$ such that the following holds $\forall g\in\{r, c\}$.
\begin{align*}
    \dfrac{1}{K}\sum_{k=0}^{K-1}J_{g, \rho}(\theta_k) = J_{g, \rho}^{\bar{\pi}}
\end{align*}
This allows us to rewrite \eqref{eq_appndx_72} as,
\begin{align}
    \label{eq_appndx_73}
    \begin{split}
        J_{r, \rho}^{\pi^*} - \mathbf{E}\big[J_{r, \rho}^{\bar{\pi}}\big] + \lambda_{\max}\mathbf{E}\big[-J_{c, \rho}^{\bar{\pi}}\big]\leq \sqrt{\epsilon_{\mathrm{bias}}} &+ f_0\dfrac{(1+\lambda_{\max})}{(1-\gamma)^2}\exp\left(-\dfrac{\mu_F}{40G^2}H\right) \\
        &+ (f_1+2)\dfrac{(1+\lambda_{\max})}{(1-\gamma)^2}\dfrac{1}{\sqrt{K}} 
    \end{split}
\end{align}

Applying Lemma \ref{lemma_appndx_constraint_extraction} and verifying (via Lemma \ref{lemma_appndx_bound_lambda}) that $\lambda_{\max}\geq 2\lambda^*$ where $\lambda^*$ is the non-negative minimizer of the dual function corresponding to the unparameterized constrained optimization \eqref{eq:original_optimization}, we can write the constraint violation rate as follows.
\begin{align}
\label{eq_appndx_74_new}
    \begin{split}
        \dfrac{1}{K}\sum_{k=0}^{K-1}\mathbf{E}\left[-J_{c, \rho}(\theta_k)\right] \leq \dfrac{2\sqrt{\epsilon_{\mathrm{bias}}}}{\lambda_{\max}} &+ 2f_0\dfrac{(1+\lambda_{\max})}{\lambda_{\max}(1-\gamma)^2}\exp\left(-\dfrac{\mu_F}{40G^2}H\right)\\ 
        &+ 2(f_1+2)\dfrac{(1+\lambda_{\max})}{\lambda_{\max}(1-\gamma)^2}\dfrac{1}{\sqrt{K}} 
    \end{split}
\end{align}

\subsection{Final Result}

Substituting $\lambda_{\max}=2/[(1-\gamma)c_{\mathrm{slater}}]\geq 2$ in \eqref{eq_appndx_67}, we get the rate of convergence of the objective as follows.
\begin{align}
        \begin{split}
        \frac{1}{K}\sum_{k=0}^{K-1}\mathbf{E}\bigg[J_{r, \rho}^{\pi^*}&-J_{r, \rho}(\theta_k)\bigg]\\
        &\leq \sqrt{\epsilon_{\mathrm{bias}}} + \dfrac{3f_0 c_{\mathrm{slater}}^{-1}}{(1-\gamma)^3}\exp\left(-\dfrac{\mu_F}{40G^2}H\right)
        + \dfrac{3(f_1+1) c_{\mathrm{slater}}^{-1}}{(1-\gamma)^3}\dfrac{1}{\sqrt{K}} 
    \end{split}
\end{align}

Similarly, we obtain the constraint violation rate as,
\begin{align}
    \begin{split}
        \dfrac{1}{K}\sum_{k=0}^{K-1}&\mathbf{E}\left[-J_{c, \rho}(\theta_k)\right] \\
        &\leq (1-\gamma)c_{\mathrm{slater}}\sqrt{\epsilon_{\mathrm{bias}}}+ \dfrac{3f_0}{(1-\gamma)^2}\exp\left(-\dfrac{\mu_F}{40G^2}H\right)
        + \dfrac{3(f_1+2)}{(1-\gamma)^2}\dfrac{1}{\sqrt{K}} 
    \end{split}
\end{align}

Let, $H$ and $K$ are chosen as follows for an arbitrary $\epsilon>0$.
\begin{align}
    \begin{split}
        H &= \dfrac{40G^2}{\mu_F}\log\left(\max\left\{\dfrac{6f_0c_{\mathrm{slater}}^{-1}}{(1-\gamma)^3}\epsilon^{-1}, \dfrac{6f_0}{(1-\gamma)^2}\epsilon^{-1} \right\}\right) = \mathcal{O}(\log(\epsilon^{-1})),\\
        K &= \max\left\{  \dfrac{36(f_1+1)^2 c_{\mathrm{slater}}^{-2}}{(1-\gamma)^6}\epsilon^{-2}, \dfrac{36(f_1+2)^2}{(1-\gamma)^4}\epsilon^{-2} \right\}
         = \mathcal{O}((1-\gamma)^{-6}\epsilon^{-2})
    \end{split}
\end{align}

For the above choice of $H$ and $K$, we have,
\begin{align}
    \label{eq_appndx_regret_constraint_epsilon}
    \begin{split}
         &\frac{1}{K}\sum_{k=0}^{K-1}\mathbf{E}\bigg[J_{r, \rho}^{\pi^*}-J_{r, \rho}(\theta_k)\bigg]\leq \sqrt{\epsilon_{\mathrm{bias}}} + \epsilon, \\
         &\mathbf{E}\left[\dfrac{1}{K}\sum_{k=0}^{K-1}-J_{c, \rho}(\theta_k)\right] \leq (1-\gamma)c_{\mathrm{slater}} \sqrt{\epsilon_{\mathrm{bias}}} + \epsilon
    \end{split}
\end{align}

Note that the expected number of steps required in Algorithm \ref{algo_sampling} is $\mathcal{O}((1-\gamma)^{-1})$. Therefore, the sample complexity required to ensure \eqref{eq_appndx_regret_constraint_epsilon} is $\mathcal{O}((1-\gamma)^{-1}HK) = \tilde{\mathcal{O}}((1-\gamma)^{-7}\epsilon^{-2})$. Finally, note that Lemma \ref{lemma_second_order} requires $H>\bar{C}\frac{G^2}{\mu_F}\log\left(\sqrt{\mathrm{d}}\frac{G^2}{\mu_F}\right)$. This can be ensured if $\epsilon$ is sufficiently small.
\section{Strong Duality and Related Lemmas}

Define the dual function associated with the unparameterized constrained optimization \eqref{eq:original_optimization} as follows.
\begin{align}
    J_{\mathrm{D}, \rho}^{\lambda} = \max_{\pi}~ \{J_{r, \rho}^{\pi}+\lambda J_{c, \rho}^{\pi}\}
\end{align}
The following lemma formally describes the strong duality result.

\begin{lemma}\citep{ding2023convergence} [Lemma 3]
    \label{lemma_appndx_strong_duality}
    If $\lambda^* \triangleq {\arg\min}_{\lambda\geq 0}J_{\mathrm{D}, \rho}^{\lambda}$ and $\pi^*$ is a solution of \eqref{eq:original_optimization}, then the following holds whenever Assumption \ref{ass_slater} is true.
    \begin{align}
        J_{r, \rho}^{\pi^*} = J_{\mathrm{D}, \rho}^{\lambda^*}
    \end{align}
\end{lemma}
It is to be mentioned that strong duality, in general, does not hold for the parameterized optimization \eqref{eq:parameterized_optimization}. The following lemma established a bound on $\lambda^*$ which becomes the foundation for choosing the value of $\lambda_{\max}$ in Algorithm \ref{algo_npg}.
\begin{lemma}\citep{ding2023convergence}[Lemma 3]
\label{lemma_appndx_bound_lambda}
    Let $\lambda^*$ be the optimal dual variable as defined in Lemma \ref{lemma_appndx_strong_duality}. The following inequalities hold where $c_{\mathrm{slater}}$ is defined in Assumption \ref{ass_slater}. 
    \begin{align*}
        0\leq \lambda^*\leq \dfrac{1}{(1-\gamma)c_{\mathrm{slater}}}
    \end{align*}
\end{lemma}
The following lemma is the main tool in decoupling the objective and constraint violation rates.
\begin{lemma}
    \label{lemma_appndx_constraint_extraction}
    Let Slater's condition (Assumption \ref{ass_slater}) hold. If $C\geq 2\lambda^*$ where $\lambda^*$ is defined in Lemma \ref{lemma_appndx_strong_duality} and $\Bar{\pi}$ is a policy such that $J_{r, \rho}^{\pi^*} - J_{r, \rho}^{\Bar{\pi}}+C[-J_{c, \rho}^{\Bar{\pi}}]\leq \zeta$ for some $\zeta>0$ then
    \begin{align}
        -J_{c, \rho}^{\bar{\pi}} \leq \dfrac{2\zeta}{C}
    \end{align}
\end{lemma}
\begin{proof}
    Define the function $v(\cdot)$ as follows.
    \begin{align}
        \label{eq_appndx_77}
        v(\tau) = \max_{\pi}\left\{J_{r, \rho}^{\pi}\big|J_{c, \rho}^{\pi}\geq \tau\right\}, ~\tau\in \mathbb{R}^{\mathrm{d}}
    \end{align}
    Let $\tau = J_{c, \rho}^{\bar{\pi}}$. Therefore, one can write the following chain of inequalities.
    \begin{align}
        \label{eq_appndx_78}
        \begin{split}
            J_{r, \rho}^{\Bar{\pi}} \overset{(a)}{\leq} v(\tau)=\max_{\pi}\left\{J_{r, \rho}^{\pi}\big|J_{c, \rho}^{\pi}\geq \tau\right\}
            &\overset{(b)}{\leq} \max_{\pi}\left\{J_{r, \rho}^{\pi}+\lambda^*(J_{c, \rho}^{\pi}-\tau)\big|J_{c, \rho}^{\pi}\geq \tau\right\}\\
            &= \max_{\pi}\left\{J_{r, \rho}^{\pi}+\lambda^*J_{c, \rho}^{\pi}\big|J_{c, \rho}^{\pi}\geq \tau\right\} - \tau\lambda^*\\
            &\overset{(c)}{\leq} J_{\mathrm{D}, \rho}^{\lambda^*} - \tau\lambda^* \overset{(d)}{=} J_{r, \rho}^{\pi^*} - \tau\lambda^*
        \end{split}
    \end{align}
    where (a) follows from the definition of $v(\tau)$, (b) uses $\lambda^*\geq 0$, (c) follows from the definition of the dual function, and (d) is a result of the strong duality (Lemma \ref{lemma_appndx_strong_duality}). Utilizing \eqref{eq_appndx_78}, we get, 
    \begin{align}
        \begin{split}
             (C-\lambda^*)(-\tau)\leq J_{r, \rho}^{\pi^*} - J_{r, \rho}^{\bar{\pi}} + C(-\tau) \leq \zeta
        \end{split}
    \end{align}
    This leads to $-\tau\leq \zeta/(C-\lambda^*)\leq 2\zeta/C$.
\end{proof}
\end{document}